%% file: main.tex
\documentclass[sigconf,screen]{acmart}

\usepackage{subcaption}
\usepackage[linesnumbered,ruled,vlined]{algorithm2e}
\usepackage{multirow}

\newcommand{\algnot}{\textnormal{\textbf{not}}\xspace}

\newcommand{\algin}{\textnormal{\textbf{in}}\xspace}

\AtBeginDocument{%
  \providecommand\BibTeX{{%
    \normalfont B\kern-0.5em{\scshape i\kern-0.25em b}\kern-0.8em\TeX}}}

\copyrightyear{2024}
\acmYear{2024}
\setcopyright{rightsretained}
\acmConference[PACT '24]{International Conference on Parallel Architectures and Compilation Techniques}{October 14--16, 2024}{Long Beach, CA, USA}
\acmBooktitle{International Conference on Parallel Architectures and Compilation Techniques (PACT '24), October 14--16, 2024, Long Beach, CA, USA}
\acmDOI{10.1145/3656019.3689611}
\acmISBN{979-8-4007-0631-8/24/10}

\begin{document}

\title{Optimizing Tensor Computation Graphs with Equality Saturation and Monte Carlo Tree Search}

\author{Jakob Hartmann}
\affiliation{%
  \institution{Department of Computer Science and Technology, University of Cambridge}
  \country{United Kingdom}}
\email{jh2422@cantab.ac.uk}

\author{Guoliang He}
\affiliation{%
  \institution{Department of Computer Science and Technology, University of Cambridge}
  \country{United Kingdom}}
\email{gh512@cam.ac.uk}

\author{Eiko Yoneki}
\affiliation{%
  \institution{Department of Computer Science and Technology, University of Cambridge}
  \country{United Kingdom}}
\email{eiko.yoneki@cl.cam.ac.uk}

\renewcommand{\shortauthors}{J. Hartmann, G. He, and E. Yoneki}

\begin{abstract}
  The real-world effectiveness of deep neural networks often depends on their latency, thereby necessitating optimization techniques that can reduce a model's inference time while preserving its performance. One popular approach is to sequentially rewrite the input computation graph into an equivalent but faster one by replacing individual subgraphs. This approach gives rise to the so-called phase-ordering problem in which the application of one rewrite rule can eliminate the possibility to apply an even better one later on. Recent work has shown that equality saturation, a technique from compiler optimization, can mitigate this issue by first building an intermediate representation (IR) that efficiently stores multiple optimized versions of the input program before extracting the best solution in a second step. In practice, however, memory constraints prevent the IR from capturing all optimized versions and thus reintroduce the phase-ordering problem in the construction phase. In this paper, we present a tensor graph rewriting approach that uses Monte Carlo tree search to build superior IRs by identifying the most promising rewrite rules. We also introduce a novel extraction algorithm that can provide fast and accurate runtime estimates of tensor programs represented in an IR. Our approach improves the inference speedup of neural networks by up to 11\% compared to existing methods.
\end{abstract}

\ccsdesc[500]{Computer systems organization~Neural networks}
\ccsdesc[500]{Computing methodologies~Optimization algorithms}
\ccsdesc[500]{Computing methodologies~Discrete space search}

\keywords{Deep Learning, Tensor Programs, Computation Graphs, Equality Saturation, Monte Carlo Tree Search, Phase-Ordering Problem}

\maketitle

\section{Introduction}
\label{sec:introduction}
Deep learning applications have achieved remarkable results in recent years in problem areas ranging from game playing \cite{alpha_zero} to computer vision \cite{stable_diffusion} and natural language processing \cite{instruct_gpt}. Many of these successes have been driven by an increase in model size, which has resulted in greater computational requirements and higher latencies. In order to use these models in practice, they undergo several optimizations steps before being deployed. One common high-level optimization technique is to transform the computation graph of the neural network into an equivalent but faster one.
\par
The traditional approach used by deep learning frameworks like TensorFlow \cite{tensorflow} and PyTorch \cite{pytorch} is to sequentially apply a set of rewrite rules, which replace individual subgraphs of the input program with optimized ones. These replacements are destructive, meaning that the original subgraphs are no longer represented in the tensor program after a rewrite rule has been applied. This phenomenon can give rise to the phase-ordering problem in which the application of one rewrite rule can eliminate the possibility to apply an even better one later on. Due to the large combinatorial search space it is often infeasible to determine the optimal ordering by brute force.
\par
Recent work \cite{tensat} has used equality saturation, a technique from compiler optimization, to address the phase-ordering problem in tensor program optimizers. Equality saturation follows a two-step process: First, an intermediate representation called equality graph (e-graph) is constructed that efficiently stores multiple optimized versions of the input program. This step is purely additive since no information is removed from the IR. The construction phase is completed once the e-graph is saturated, i.e. the application of rewrite rules no longer adds any information to the e-graph, or when a time or memory limit has been reached. Then, in the second step, an extraction algorithm is used to obtain the optimal input program from the e-graph.
\par
Equality saturation solves the phase-ordering problem in situations where the e-graph can saturate and represent all possible versions of the input program. In practice, however, this rarely occurs because the e-graph tends to explode rapidly and reach a memory limit before it has saturated. In these cases, the quality of the e-graph and the final solution depend on which rewrite rules were applied before the memory limit was reached. Thus, the phase-ordering problem is reintroduced into the construction phase of equality saturation when the e-graph cannot saturate. To address this issue, we use Monte Carlo tree search (MCTS) to identify the most promising rewrite rules in the e-graph construction phase.
\par
A second issue that limits the effectiveness of equality saturation is the dependency on a good extraction algorithm. There are two types of extraction algorithm: Integer Linear Programs (ILPs) and greedy extractors. While ILPs are guaranteed to find the optimal solution, their search time scales exponentially with the e-graph size, making them impractical for many problem settings. Greedy extractors on the other hand are fast, but often do not find the optimal solution, because they fail to take common subexpressions into account. Prior work \cite{tensat} has shown that they can even extract tensor programs that are slower than the original one. 
To address this issue, we propose a novel cost function that allows greedy extractors to handle common subexpressions.
\par
Specifically, we make the following contributions:
\begin{enumerate}
    \item We develop an equality saturation-based tensor program optimizer that uses Monte Carlo tree search to construct the equality graph. We show that our approach improves the inference speedup of neural networks by up to 11\% compared to existing methods.
    \item We devise a new greedy extraction algorithm that retrieves superior tensor programs from an equality graph by taking common subexpressions into account. We show that our algorithm provides MCTS with an accurate and fast reward signal during the construction phase.
\end{enumerate}

\section{Background}
\label{sec:background}

\subsection{Term Rewriting Systems}
\label{sec:background_trs}
The optimization problem of transforming the computation graph of a neural network into an equivalent but faster one can be described as a term rewriting system (TRS). Following the notation of Klop 1993 \cite{term_rewriting_systems}, a TRS is a tuple $(\Sigma, R)$ where $\Sigma$ is an alphabet and $R$ is a set of rewrite rules. Terms (or expressions) can be defined recursively using the variables, constants, and function symbols $F$ given by the alphabet. The set of terms over the alphabet $\Sigma$ is denoted as $\text{Ter}(\Sigma)$. Each rewrite rule $r: t \rightarrow s$ reduces a term $t$ from the alphabet into another term $s$. At each step in the optimization procedure, applying a rewrite rule $r$ leads to a set of rewrites $\sigma(t) \rightarrow_r \sigma(s)$ for substitutions $\sigma$. Substitutions describe a mapping $\text{Ter}(\Sigma) \rightarrow \text{Ter}(\Sigma)$ where $\sigma(F(t_1, ..., t_m)) = F(\sigma(t_1), ..., \sigma(t_m))$. By sequentially performing $n$ rewrites, $t_0$ is reduced to $t_n$. 
\par
We can say that the application of rewrite rule $r$ at time step $i \in \mathbb{N}$ gives rise to the phase-ordering problem if it eliminates the possibility to apply a more favourable rewrite rule $r'$ at time step $j \in \mathbb{N^*}$ where $j > i$. The prototypical example that is often used to illustrate the destructive nature of traditional term rewriting systems is the expression $(a * 2) / 2$ \cite{egg, mcts_geb}. Assuming we have a set of rewrite rules $R = \{x * 2 \rightarrow x \ll 1; (x * y) / z \rightarrow x * (y / z); x/x \rightarrow 1 (x \neq 0); x * 1 \rightarrow x\}$, we can apply the strength-reduction operation $x*2 \rightarrow x \ll 1$ to replace the expensive multiplication instruction $a * 2$ with the cheaper bitshift instruction $a \ll 1$. However, this eliminates the future possibility of canceling out the fraction altogether to eventually arrive at the optimal solution $a$.

\subsection{Equality Saturation}
\label{sec:background_es}
Motivated by the phase-ordering problem in traditional compilers, Tate et al., 2009 \cite{equality_saturation} proposed the equality saturation framework. Instead of destructively modifying the input program, equality saturation uses an e-graph as an IR to efficiently store multiple different versions of the input program. First, the e-graph is constructed by iteratively applying all rewrite rules before the optimal solution is extracted in a second step. An e-graph consists of equivalence classes (e-classes) and equivalence nodes (e-nodes) that are used to store congruence relations over different terms. An e-class is a set of e-nodes that represent equivalent terms. E-nodes are variables, constants, and function symbols from the underlying alphabet and can have an arbitrary number of children e-classes associated with them. Congruence relations are equivalence relations that are preserved by the application of rewrite rules. Two terms are congruent to each other if applying the same set of rewrite rules results in equivalent terms.

\begin{figure}[h]
    \captionsetup[subfigure]{justification=centering}
     \raggedright
     \begin{subfigure}[t]{0.19\textwidth}
         \centering
         \includegraphics[height=3.7cm, keepaspectratio]{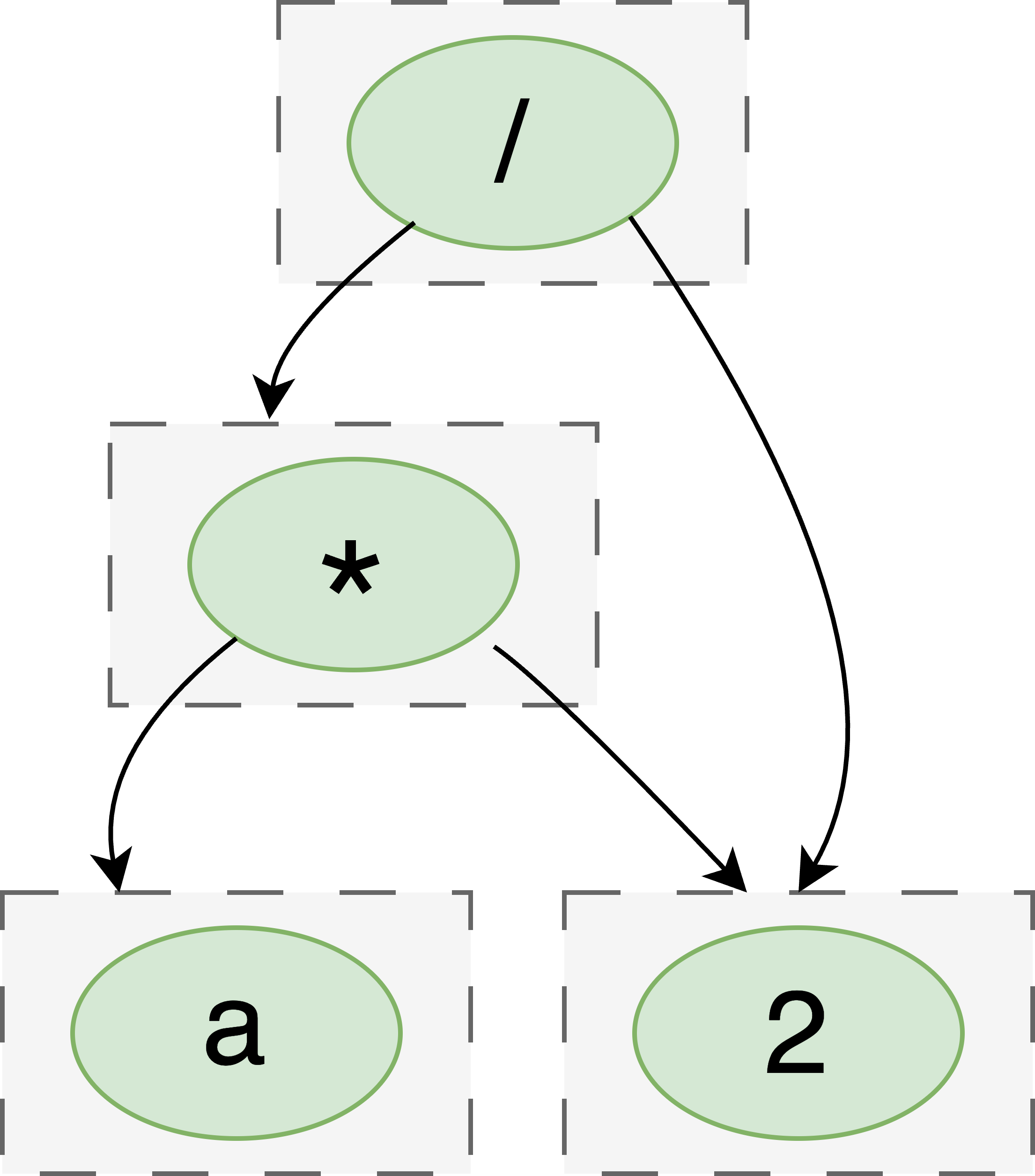}
         \caption{Initial e-graph}
         \label{fig:egraph_example_start}
     \end{subfigure}
     \begin{subfigure}[t]{0.19\textwidth}
         \centering
         \includegraphics[height=3.7cm, keepaspectratio]{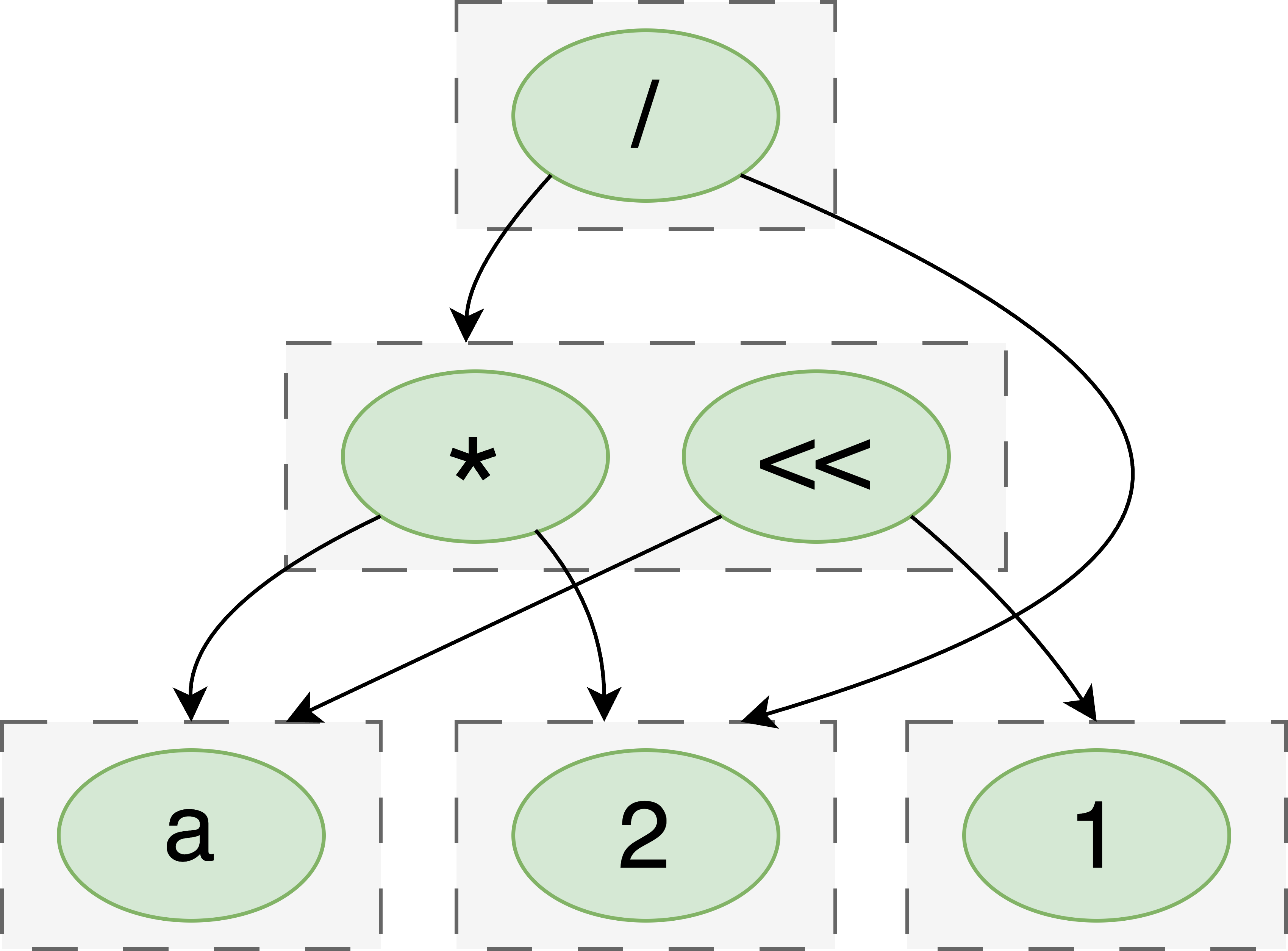}
         \caption{E-graph after applying \\ $x * 2 \rightarrow x \ll 1$}
         \label{fig:egraph_example_bitshift}
     \end{subfigure}
        \caption{Example e-graphs for expression $(a * 2) / 2$. E-classes are represented as rectangles, e-nodes are shown in circles.}
        \label{fig:egraph_examples}
\end{figure}

\subsubsection{Construction}
\label{sec:background_es_construction}
In the beginning, the e-graph is initialized with the input program. Each e-node represents one variable, constant, or function symbol from the alphabet. At the start, each e-class consists of exactly one e-node. Figure \ref{fig:egraph_example_start} shows the initial e-graph corresponding to the expression $(a * 2) / 2$. The dashed rectangles represent e-classes, the solid circles represent e-nodes and the parent-child relationships are depicted by arrows. After initialization, the algorithm iterates over all rewrite rules and searches for the left-hand side of each rule in the e-graph. If the pattern is found, the e-nodes corresponding to the right-hand side are inserted and merged with the respective e-classes. Compared to the traditional approach, this process is purely additive and the left-hand side pattern will remain in the e-graph. 
\par
Figure \ref{fig:egraph_example_bitshift} illustrates this concept using the previous example. After applying the rewrite rule $x * 2 \rightarrow x \ll 1$, the e-graph encodes two programs: $(a * 2) / 2$ and $(a \ll 1) / 2$. Thus, the rewrite has not destroyed any information and it is still possible to cancel out the fraction and obtain the optimal solution $a$ in future iterations.
\par
The construction phase is completed once the e-graph has saturated, i.e. when no rewrite rule can add any further information to the e-graph, or when a given time or memory limit has been reached. In the first case, the e-graph represents all possible versions of the input program based on the provided rule set. Thus, the phase-ordering problem is not an issue. In the latter case, however, the e-graph does not encode all possible versions and therefore the quality of the e-graph and the extracted solution depend on the rewrite rules that have been applied up to that point. In these situations the phase-ordering problem is reintroduced into the construction phase.

\begin{figure*}[h]
     \centering
     \begin{subfigure}[t]{0.48\textwidth}
         \centering
         \includegraphics[width=\textwidth]{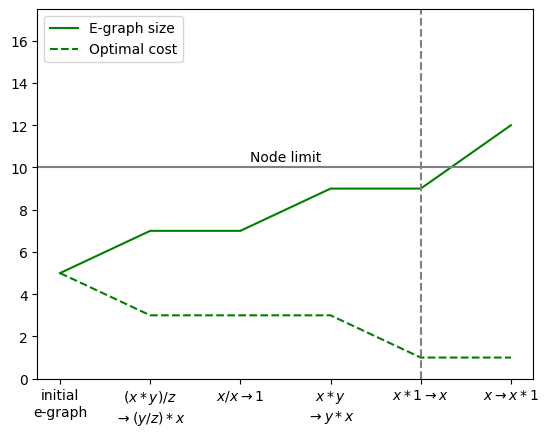}
         \caption{Good rule ordering. The optimal expression $a$ is found before the e-graph reaches the node limit.}
         \label{fig:toy_example_good_ordering}
     \end{subfigure}
     \hfill
     \begin{subfigure}[t]{0.48\textwidth}
         \centering
         \includegraphics[width=\textwidth]{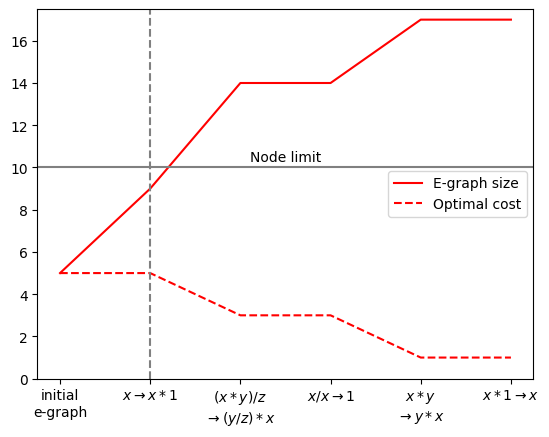}
         \caption{Bad rule ordering. The optimal solution is not found within the node limit.}
         \label{fig:toy_example_bad_ordering}
     \end{subfigure}
        \caption{Simple example of the phase-ordering problem during e-graph construction. The input expression is $a * 2 / 2$, the node limit is set to 10, and the cost is calculated based on the Abstract Syntax Tree size of the extracted expression. The x-axis shows the rewrite rules being applied and the y-axis displays the associated e-graph size together with the optimal cost at each point.}
        \label{fig:toy_example}
\end{figure*}

Figure \ref{fig:toy_example} shows a simple example of this case based on the expression $(a * 2) / 2$. The rewrite rule applications are plotted on the x-axis, and the y-axis represents the e-graph size (solid line) and the optimal cost (dotted line). The latter is calculated based on the Abstract Syntax Tree (AST) size of the optimal expression extracted from the respective e-graph. In this example, we assume a hypothetical memory/node limit of 10, meaning that we would stop the e-graph construction once a rewrite rule application results in an e-graph that has 10 or more e-nodes. 

In Figure \ref{fig:toy_example_good_ordering}, the rewrite rules are applied in the ideal order, enabling the e-graph to capture the optimal solution $a$ with an e-graph size of 9, i.e. within the node limit. In Figure \ref{fig:toy_example_bad_ordering}, the same rewrite rules are applied in a different, non-ideal order, requiring an e-graph size of 17 to capture the optimal solution. If the e-graph construction had been stopped after reaching the node limit, the optimal solution would not have been found.  In practice, the problem is aggravated by the fact that some rules lead to an exponential explosion of the e-graph, making it even harder to find the optimal solution. This example illustrates the reintroduction of the phase-ordering problem in equality saturation and motivates a solution like ours, which identifies the best rewrite rules during e-graph construction.

\subsubsection{Extraction}
\label{sec:background_es_extraction}
After completing the first phase, the optimal solution needs to be extracted from the e-graph. This process relies on a cost model that can rank the encoded programs with regards to the optimization objective. For basic tasks like simplifying mathematical expressions, the AST size can be used. For more complex tasks like optimizing the computation graphs of neural networks, more sophisticated methods are needed. We will discuss our choice of cost model in Section \ref{sec:methodology_implementation}. Based on this model, an ILP or greedy extractor can then be used to obtain the optimized program.

\paragraph{Integer Linear Programs}
Following the notation from \cite{tensat}, the ILP for extracting the optimal tensor program from an e-graph can be formulated as:

\begin{align}
    \text{minimize} \quad & \sum_{i} c_i x_i \tag{1a} \label{ilp:objective} \\
    \text{subject to} \quad & x_i \in \{0,1\} \tag{1b} \label{ilp:binary_constraint} \\
    & \sum_{i \in e_0} x_i = 1 \tag{1c} \label{ilp:root_constraint} \\
    & \forall i, \forall m \in h_i, \ x_i \leq \sum_{j \in e_m} x_j \tag{1d} \label{ilp:child_constraint} \\
    & \forall i \in l, \ x_i = 0 \tag{1e} \label{ilp:cycle_constraint}
\end{align}

Where $i$ is an e-node, $c_i$ its cost as determined by the cost model, and $h_i$ the set of children e-classes. $m$ denotes an e-class and $e_m$ the set of e-nodes of that e-class. The objective of the ILP is to extract a valid program with the lowest overall cost. Constraint \ref{ilp:binary_constraint} defines $x_i$ as a binary variable, which encodes whether the respective e-node is selected or not. Constraint \ref{ilp:root_constraint} asserts that one e-node in the root e-class $(m = 0)$ needs to be part of the final program. Constraint \ref{ilp:child_constraint} ensures that if an e-node is selected, so are its children e-classes. And the final constraint \ref{ilp:cycle_constraint} restricts the solution to all e-nodes that are not part of some blacklist $l$. The latter is required to ensure that the extracted program is a directed acyclic graph.

\paragraph{Greedy Extractors}
Greedy extractors as used in \cite{egg, tensat} follow a bottom-up approach, starting at the e-graph's leaf nodes working their way up to the root node. For each e-class, they iterate over all e-nodes and call a cost function which calculates the sum of the e-node's operator cost (as determined by the cost function) and the costs of all children e-classes. Afterwards, the lowest cost together with its associated e-node is saved as the reference cost for that e-class. This process is repeated until the costs of all e-classes have converged. The overall cost of the best tensor program encoded in the e-graph then corresponds to the cost of the root e-class. The associated computation graph can be constructed by starting at the root e-class and recursively selecting the best e-node in each e-class before proceeding to the e-node's children.
\par
Greedy extractors are significantly faster than ILPs and scale well to larger e-graph sizes. However, they are not guaranteed to find the optimal solution if the program contains common subexpressions. This is a significant issue in the case of tensor computation graphs, because any multi-input node (e.g. a skip connection) will create shared subgraphs. These subgraphs will then be considered multiple times by the cost function, leading to an overestimate of the real latency. This inaccuracy, in turn, can lead to the extraction of suboptimal tensor programs. Yang et al., 2021 \cite{tensat} have shown that in some cases, the "optimized" computation graph can even be slower than the original one. In Section \ref{sec:methodology_extraction} we will analyze this problem in more detail and introduce our improved cost function.

\subsection{Monte Carlo Tree Search}
\label{sec:background_mcts}
MCTS is a model-based planning algorithm originally developed for the use in computer Go \cite{bandit_based_monte_carlo_planning, efficient_selectivity_and_backprop_operators_in_mcts, mogo}. The main idea is to build a search tree by balancing the exploitation of states that have led to high rewards in the past with the exploration of new ones. Each node in the tree represents a state and each edge corresponds to an action. MTCS iteratively works through four steps (selection, expansion, simulation, update) to grow the search tree into the most promising areas of the search space. It terminates once a pre-defined time or iteration limit has been reached. At the end of the search, the best action is determined based on the root's child node with the highest visit count or highest average value. We will discuss how we use MCTS to construct the e-graph in Section \ref{sec:methodology_construction}.

\section{Related Work}
\label{sec:related_work}

\subsection{Equality Saturation}
\label{sec:related_work_equality_saturation}
Although the theoretical foundations for the equality saturation approach were laid by Tate et al. in 2009 \cite{equality_saturation}, the practical applicability has long been hampered by the necessity to develop domain-specific implementations for each use-case. This gap was closed by Willsey et al., 2021 \cite{egg} with the e-graphs good (egg) library, which allows users to define their own alphabets and rewrite rules on top of a generic equality saturation framework. egg has been used in a variety of projects, for example, to optimize floating point expressions \cite{egg} and for numerical hardware design \cite{automating_constraint_aware_datapath_optimization_using_egraphs}.

\subsection{Tensor Program Optimization}
\label{sec:related_work_tensor_program_optimization}
Rewriting computation graphs of neural networks requires a good set of rewrite rules. Traditionally, human experts hand-craft these rules by identifying non-optimal source graphs and match them with equivalent but optimized target graphs. TASO \cite{taso} automates this process by generating all possible substitution candidates up to a certain size and validating them against human provided operator specifications. Afterwards, it applies MetaFlow's \cite{metaflow} cost-based backtracking search to jointly optimize graph substitutions and data layouts. The cost model measures the runtime of individual operators on the underlying hardware and then calculates their sum to obtain the overall runtime estimate of the tensor program.

\paragraph{Equality saturation} To address the phase-ordering problem of traditional graph rewriting approaches, Yang et al., 2021 \cite{tensat} introduce a tensor program optimizer called TENSAT based on equality saturation. The authors build on TASO and replace its backtracking search with a two-step e-graph construction and extraction process. To benefit from all rewrite rules generated by TASO, Yang et al. extend the construction phase to support multi-pattern rewrite rules. Multi-pattern rewrite rules consist of two source patterns, both of which need to be present in the e-graph for the target patterns to be applied.
\par
Although TENSAT showed significant improvements over TASO in terms of optimization results and times, it has several shortcomings. First, all rewrite rules are applied sequentially during the construction phase, thereby leading to sub-optimal results if the e-graph hits a memory limit. Second, since the multi-pattern rewrite rules tend to rapidly explode the e-graph, the authors had to limit their application to one or two iterations, thus also restricting their effectiveness. Third, due to the unreliability of greedy extractors, TENSAT uses ILPs for the extraction step. This, however, restricts its application to smaller e-graphs. These three problems emphasise the significance of the phase-ordering problem in equality saturation as well as that of reliable greedy extractors and are addressed by our work.

\paragraph{Deep Reinforcement Learning}
He et al., 2023 \cite{xrlflow} also build on TASO, but replace the cost-based backtracking search with deep reinforcement learning. Their model-free RL agent receives the encoded tensor program in form of a graph neural network (GNN) as input and sequentially decides which rewrite rule to apply next. The authors show that their approach X-RLflow outperforms TASO due to the agent's ability to trade-off short-term performance losses in favour of long-term runtime reductions. However, these improvements come at the cost of an extensive pre-training phase. Our approach on the other hand is planning-based and can optimize tensor programs wtihout prior training.
\par
\paragraph{ML compilers}
There are several other ML compilers that focus on different aspects of the optimization routine: Hidet \cite{hidet} uses a task-based programming paradigm to embed the scheduling process into tensor programs. TACO \cite{taco} and SparseTIR \cite{sparsetir} optimize compound tensor algebra expressions consisting of sparse and dense tensors. Ansor \cite{ansor} employs a task scheduler, program sampler, and performance tuner to itereratively optimize graph partitions. TVM \cite{tvm} is an end-to-end deep learning compiler that takes high-level representations of neural networks and maps them to low-level optimized code. ONNX Runtime \cite{onnx} is an inference engine that enables interoperability between different machine learning frameworks and provides support for model optimizations such as quantization and model pruning. In comparison to these compilers, we focus on the high-level rewriting of the tensor computation graph.

\subsection{Monte Carlo Tree Search}
\label{sec:related_work_mcts}

\begin{figure*}[!h]
\includegraphics[width=\linewidth]{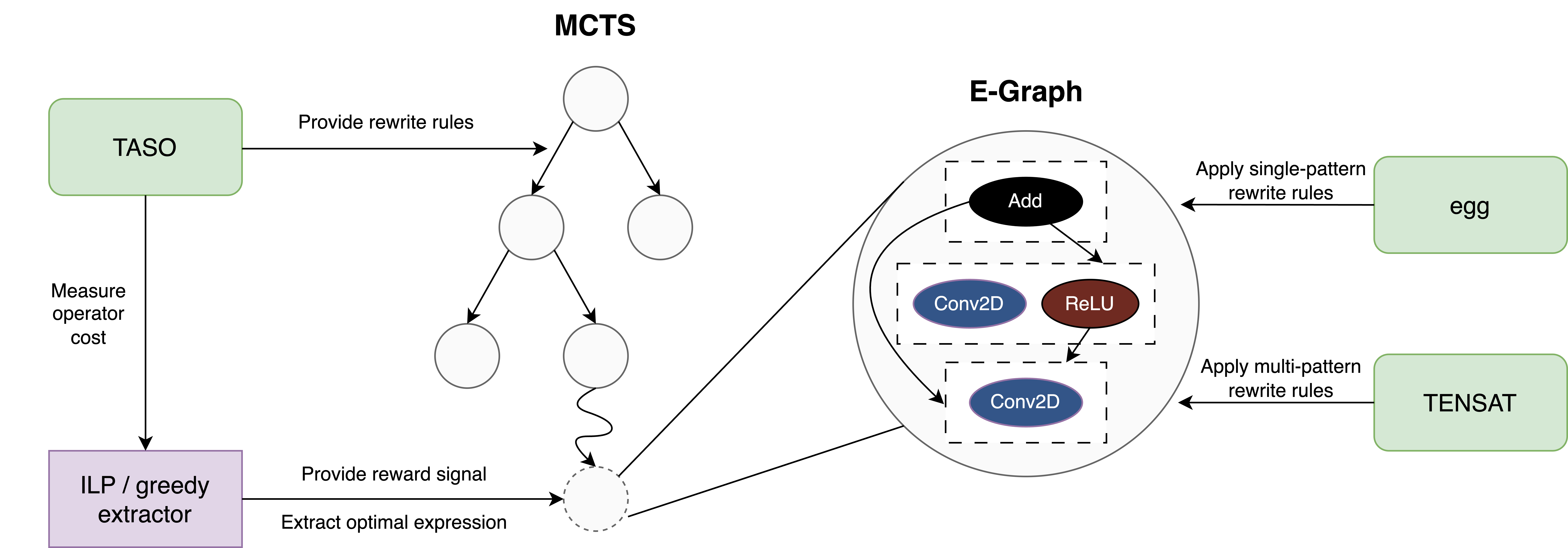}
    \caption{Overview of our tensor program optimizer using equality saturation and MCTS}
    \label{fig:tensor_overview}
\end{figure*}

He et al., 2023 \cite{mcts_geb} also use MCTS to address the phase-ordering problem in equality saturation. Their approach MCTS-GEB (MCTS is a Good E-graph Builder) decides which rewrite rules to apply during e-graph construction and the authors show that they can find expressions with up to 49x lower cost compared to egg. However, their work is limited in multiple ways, as the application is restricted to a synthetic benchmark suite consisting of randomly generated toy expressions from two test domains. The real-world use case of optimizing tensor programs is significantly more complex, thereby necessitating several changes to the underlying approach. 
\par
First, the reward signal for MCTS can no longer be obtained by a traditional greedy extractor, since the structural dependencies of tensor computation graphs need to be taken into account to obtain accurate runtime estimates. Second, the AST size cost function needs to be replaced by a domain-specific one that accounts for the different runtimes of each tensor operation. Third, the significantly larger action space, the existence of multi-pattern rewrite rules and the need to avoid cycles in the e-graph pose additional problems which we will address in the following section.

\section{Methodology}
\label{sec:methodology}
In this section, we will introduce our tensor graph rewriting approach based on equality saturation and Monte Carlo tree search. We will start with the e-graph construction phase and explain how MCTS can mitigate the phase-ordering problem by predicting which rewrite rules to apply. We will then move on to the e-graph extraction phase, where we analyze the shortcomings of existing methods and introduce our own approach. Finally, we will give a brief overview of our open-source implementation.

\subsection{E-Graph Construction}
\label{sec:methodology_construction}
A high-level overview of the optimization procedure is shown in Algorithm \ref{alg:high_level_overview}. To start with, the e-graph is initialized with the input tensor program. In this e-graph, e-nodes correspond to tensors (e.g. weights, inputs) and tensor operations (e.g. ReLU, convolutions), e-classes represent equivalent tensors / tensor operations, and the parent-child relationships between e-nodes and e-classes correspond to the flow of tensors. After initialization, the e-graph is constructed sequentially by running MCTS, applying the best rewrite rule based on the results, and then repeating the process until the e-graph has either saturated or reached a memory limit. At the end, the best tensor program is extracted from the e-graph.

\begin{algorithm}[h]
\caption{Tensor program optimization}
\label{alg:high_level_overview}

\DontPrintSemicolon
\KwIn{computation graph $G$, rules $R$, search budget $N$}
\KwOut{optimized computation graph}

$egraph \gets $ \text{initialize\_egraph}($G$)\;
\While{\algnot $egraph$.\textnormal{saturated\_or\_reached\_limit()}}{
  $root \gets $ create\_node($egraph$, None) \tcp*{start MCTS}
  \For{$i \gets 1$ \KwTo $N$}{
    $node \gets $ select($root$)\;
    $childNode \gets $ expand($node$, $R$)\;
    $reward \gets $ simulate($childNode$)\;
    update($childNode$, $reward$)\;
  }
  $egraph$.apply($root$.best\_child()) \tcp*{end MCTS}
}
\Return{$egraph$.\textnormal{extract()}}\;
\end{algorithm}

Each MCTS search iteratively builds a search tree based on the current e-graph. In our setting, this e-graph corresponds to the root node of the search tree, edges represent rewrite rules, and child nodes correspond to the e-graph after the rewrite rule has been applied. In addition to the value $v$ and visit count $n$, each node stores a boolean $s$, indicating whether the node has saturated or not, and a blacklist $b$, which keeps track of all rewrite rules that we know would lead to saturated child nodes. A saturated node is one where the rewrite rule leading up to it has not resulted in any changes to the e-graph, i.e. the node's e-graph is identical with its parent node's e-graph. By default, most rewrite rules will not add any information to the e-graph, because rules may only be applicable to specific neural network architectures or need to be enabled first by other rewrite rules. We prune all nodes that have already saturated as well as all rewrite rules that we know will not change the e-graph. To build the search tree, MCTS iteratively works through four stages:

\begin{figure*}[h]
     \raggedright
     \begin{subfigure}[t]{0.25\textwidth}
         \centering
         \includegraphics[height=8.25cm, keepaspectratio]{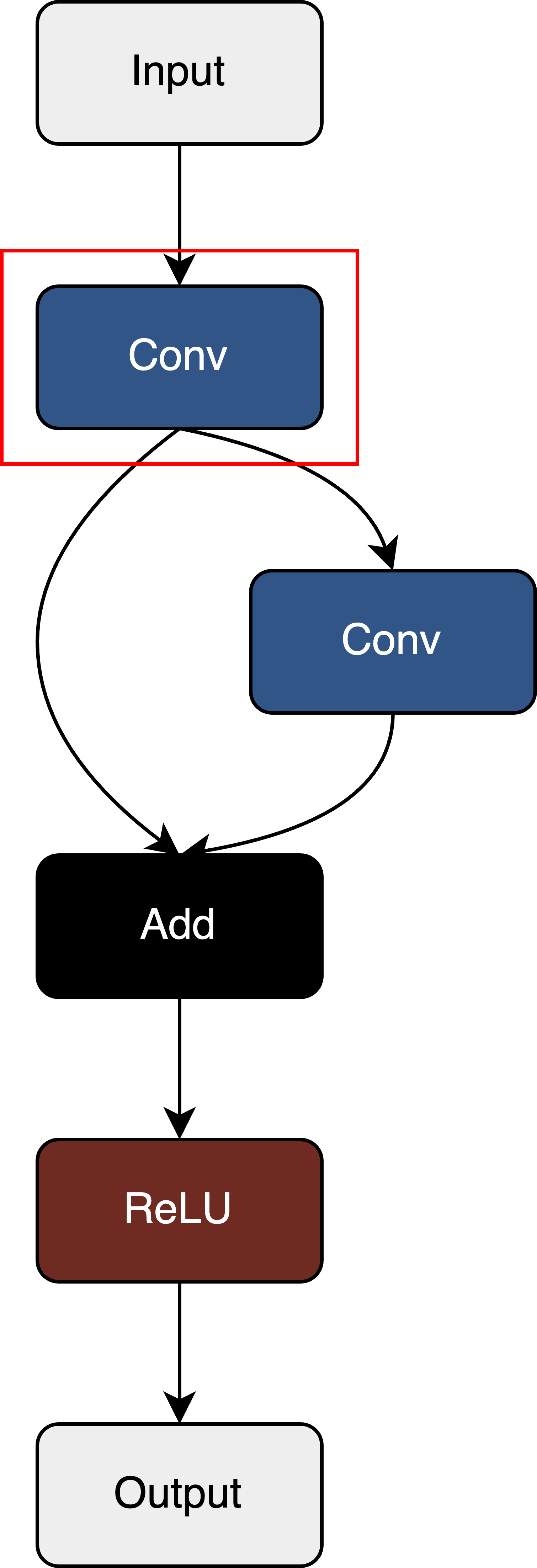}
         \caption{Tensor graph}
         \label{fig:resblock_annotated}
     \end{subfigure}
     \begin{subfigure}[t]{0.25\textwidth}
         \centering
         \includegraphics[height=8.25cm, keepaspectratio]{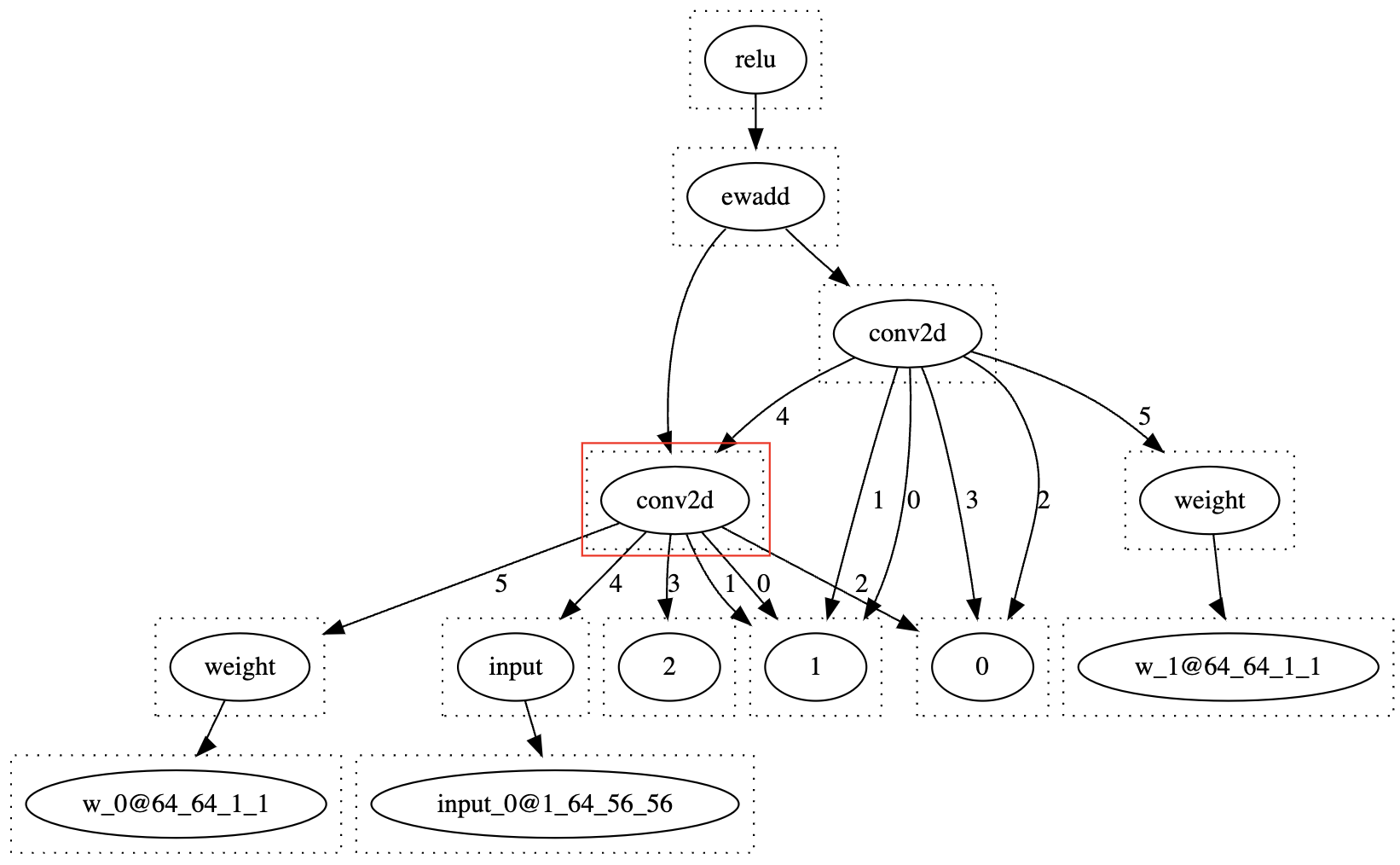}
         \caption{E-graph}
         \label{fig:resblock_e_graph}
     \end{subfigure}
        \caption{Simple example of a neural network in which greedy extractors with existing cost functions overestimate the true graph runtime. The convolution operation marked in red is a common subexpression and thus counted twice, once by the add operation and once by the second convolution operation.}
        \label{fig:resblock_example}
\end{figure*}

\paragraph{Selection} Starting at the root node, MCTS traverses the search tree. To ensure that the tree grows in both width and depth, the traversal stops at each node with 50\% probability - we adopt this hyperparameter setting from \cite{mcts_geb} - to proceed to the expansion phase. In the other 50\% of cases, one of the non-saturated child nodes is selected. To balance the exploitation and exploitation of states, the selection is based on the nodes' UCB1 (Upper Confidence Bounds) values \cite{ucb1}:

\setcounter{equation}{1}

\begin{equation}
    UCB1 = \frac{v}{n} + c * \sqrt{\frac{\ln N}{n}}
\end{equation}
where $c$ is an exploration constant and $N$ is the visit count of the parent node. The first term is responsible for the exploitation of rewrite rules that have led to high runtime reductions in the past, and the second term is the exploration term and favours rewrite rules that have previously been rarely applied.

\paragraph{Expansion} Once the tree traversal has stopped or reached a leaf node, a random rewrite rule which is not on the node's blacklist is sampled and applied to the e-graph. The resulting e-graph is then used to initialize the new child node. If the old and new e-graph are identical, the saturation flag is set to true and the node will not be visited in future iterations. If the node has not saturated, we apply a post-processing step to the new e-graph to immediately prune all rewrite rules that cannot change the e-graph in future iterations.
\par
To this end, we iterate over all single-pattern rewrite rules and check whether the respective source pattern is represented in the e-graph. If not, the rule is added to the node's blacklist and cannot be selected in future iterations. For multi-pattern rewrite rules, it is sufficient if one source pattern is not represented in the e-graph for the rule to be added to the blacklist. This approach can prune most non-applicable rewrite rules, but not all. In some cases, the source pattern is represented in the e-graph but the target pattern cannot be inserted as this would introduce a cycle in the e-graph. We disregard these cases in our pruning approach and instead adopt TENSAT's cycle filtering approach \cite{tensat} to ensure that the final tensor program is a directed acyclic graph.

\paragraph{Simulation} During simulation, we randomly select rewrite rules and apply them to the e-graph. We favour random over heavy rollouts, as prior research has shown that low-bias, high-variance strategies are often superior to high-bias and low-variance ones \cite{analysis_of_mcts}. We stop after a pre-defined number of simulation steps or if the e-graph has saturated or hit a memory limit. After each simulation step, we extract the best tensor program from the e-graph to calculate the runtime improvement over the previous one. The simulation reward is then the sum of all runtime improvements:

\begin{equation}
    r = \sum_{i=1}^{maxSimSteps} max(runtime_{i-1} - runtime_{i}, 0)
\end{equation}

We introduce the $\max$ operator to prevent the reward from turning negative in rare situations where the runtime increases due to the stochasticity of the cost model (see Section \ref{sec:methodology_implementation}) or the non-optimality of the greedy extractor. 

\paragraph{Update} In the final step, the simulation reward $r$ is backpropagated from the child to the root node. The statistics of each node are updated as follows: $v = v + r$ and $n = n + 1$.

\input{res/tables/extraction_methods_base_cost_comparison_a100}

By iterating through these four steps repeatedly, the node statistics will converge to their true underlying values and MCTS will focus on the most promising areas of the search space. Once the search budget has been exhausted, the best rewrite rule is selected based on the root's child node with the highest average value. This rewrite rule is then applied to the original e-graph, which serves as the root node in the next iteration. To reduce the optimization time, it is possible to reuse the subtree below the best child node as the starting point for the next iteration.

\subsection{E-Graph Extraction}
\label{sec:methodology_extraction}

Most equality saturation-based applications perform a single extraction at the end of the e-graph construction phase. In our setting, however, we also need to perform multiple extractions during each simulation step to obtain the necessary reward signal for MCTS. Therefore, we cannot tolerate the long optimization times of ILPs and instead have to rely on greedy extractors as an alternative. In addition to being fast to obtain, we also require the extraction results to be an accurate reflection of the optimal tensor program represented in the e-graph to ensure that the search tree grows in the most promising areas of the search space.

\paragraph{Problem} Tables \ref{tab:comparison_extraction_methods} and \ref{tab:comparison_extraction_methods_p100} show that greedy extractors which use existing cost functions fail to do so and significantly overestimate the true graph runtimes. The only neural network architecture for which the prediction matches the ILP estimate is VGG-19 \cite{vgg}. Wrong runtime estimates are not a problem by themselves as long as the relative ordering between different tensor programs represented in the e-graph is preserved. However, Yang et al., 2021 \cite{tensat} have shown that this is not the case and that in some instances the extracted program can be even slower than the original one.  This finding not only poses a problem because it conflicts with the optimization objective, but also challenges the fundamental idea of equality saturation whereby the original tensor program will always be encoded and should thus be recoverable from the e-graph.

\paragraph{Analysis} Comparing VGG-19 with the other model architectures shows that it has the only computation graph without multi-input nodes and shared subgraphs. To illustrate how these can give rise to inaccurate predictions, Figure \ref{fig:resblock_example} provides a simple example based on a skip connection. Figure \ref{fig:resblock_annotated} shows the computation graph of a residual block and Figure \ref{fig:resblock_e_graph} the corresponding e-graph. Existing cost functions such as the ones used in egg and TENSAT, determine the cost of an e-node by adding its operator cost to the sum of all its children e-classes. If we use this approach to calculate the cost of the \textit{Add} node in the example, we will sum over both \textit{Conv} nodes. However, since the cost of the first convolution operation marked in red is already included in the second one, we would overestimate the true graph runtime. If several such blocks are then stacked on top of each other to form a residual network, the errors add up exponentially. An intuitive approach to solve this problem would be to store centrally which e-classes have already been counted and not consider them a second time. While this results in correct estimates for initial e-graphs where each e-class contains exactly one e-node (e.g. Figure \ref{fig:resblock_e_graph}), it fails once the e-graph grows. Therefore, we need a more sophisticated approach to deal with cost explosions caused by multi-input nodes and shared subgraphs.

\begin{algorithm}[h]
\caption{Our e-node cost function}
\label{alg:cost_function}

\DontPrintSemicolon

\KwData{$eclassHist, bestEnodeCost, bestEnodeHist$}
\KwIn{$enode, eclass, prevEclass, costs$}
\KwOut{$enodeCost$}

\tcc{update data if e-class has changed}
\If{$prevEclass \neq eclass$}{
    $eclassHist[prevEclass] = bestEnodeHist$\;
    $bestEnodeCost = \infty$\;
}

\tcc{calculate e-node cost}
$enodeHist = \{\}$\;
$enodeCost = $ GetOperatorCost($enode$)\;

\For{$child$ \algin $enode$.\textnormal{children()}}{
    $childCost = 0$\;
    \uIf(\tcp*[f]{case 1}){$child$ \algin $enodeHist$}{
        \textbf{continue}\;
    }
    \uElseIf(\tcp*[f]{case 2}){$child$ \algin $eclassHist$}{
        $maxCost = 0$\;
        \For{($key$, $value$) \algin $eclassHist[child]$}{
            \If{$key$ \algin $enodeHist$}{
                $maxCost = $ max($maxCost, value$)\;
            }
            \Else{
                $enodeHist[key] = value$\;
            }
        }
        $childCost = $ max($costs[child]$ - $maxCost, 0$)\;
    }
    \Else(\tcp*[f]{case 3}){
        $childCost = costs[child]$\;
    }
    $enodeHist[child] = childCost$\;
    $enodeCost = enodeCost + childCost$\;
}

\tcc{update data if cheaper e-node was found}
\If{$enodeCost < bestEnodeCost$}{
    $bestEnodeHist = enodeHist$\;
    $bestEnodeCost = enodeCost$\;
}

\Return{$enodeCost$}\;

\end{algorithm}

\paragraph{Solution} The main idea behind our proposed solution is to keep track of the constituent costs of each e-class and e-node to prevent counting shared subgraphs multiple times. The constituent costs can be seen as an e-class'/e-node's history indicating which e-classes have contributed to its current cost. Algorithm \ref{alg:cost_function} shows the pseudoalgorithm for our improved cost function. To calculate the cost of an e-node, the function iterates over all children e-classes and considers three possible scenarios:
\begin{enumerate}
    \item If the e-class is already included in the e-node's history (i.e. in the constituent costs of any of its children e-classes), it is ignored. In the example from Figure \ref{fig:resblock_example}, the cost function would skip the \textit{Conv} node marked in red if it had already iterated over the other \textit{Conv} node.
    \item If the e-class itself is not included in the e-node's history, but its constituent costs overlap with those of other children e-classes, only the non-overlapping ones are added to the e-node's cost. In our example, this scenario would occur if the cost function first iterates over the \textit{Conv} node marked in red. Then, only the operator cost of the second \textit{Conv} node would be added, but none of its constituent costs. 
    \item Else, the full cost of the child e-class is added to the e-node's cost.
\end{enumerate}

In scenarios 2) and 3), the child e-class and all its non-overlapping constituent costs are added to the e-node's history. If, at the end, the e-node's final cost is lower than that of all other e-nodes in its e-class, the e-class' constituent costs are updated with the e-node's history. This ensures that the e-class' cost and constituent costs always correspond to the best e-node and allows the cost function to ignore subgraphs that have already been considered.

\paragraph{Results} The last line in Tables \ref{tab:comparison_extraction_methods} and \ref{tab:comparison_extraction_methods_p100} shows that our cost function enables greedy extractors to match the accuracy of ILPs for the initial e-graphs of all architectures except NasNet-A. NasNet-A \cite{nasnet} is a special type of model, as it was artificially generated using neural architecture search (NAS). NAS can produce nested structures which in rare circumstances result in overlapping constituent costs not being treated 100\% correctly. An illustration of this problem, which was derived from the NasNet-A computation graph, is provided in Figure \ref{fig:failure_case} in the supplementary material. Nevertheless, our runtime estimates for NasNet-A are orders of magnitude more accurate than ones produced by existing cost functions. It is important to note that although we have focused our attention on tensor programs, our cost function can improve the performance of greedy extractors for all programs with common subexpressions. In Section \ref{sec:experiments}, we will analyse how the improved prediction accuracy affects downstream performance.

\subsection{Implementation}
\label{sec:methodology_implementation}

We built our tensor program optimizer on top of MCTS-GEB, egg, TASO and TENSAT. A high-level overview of our open-source implementation is shown in Figure \ref{fig:tensor_overview}. The optimizer receives the original tensor program as input and initializes the e-graph. The single- and multi-pattern rewrite rules are provided by TASO. In each iteration, MCTS initializes the root node with the current e-graph and searches for the best rewrite rule to apply. The reward signal during the simulation phase is obtained by extracting the best tensor program from the node's e-graph and calculating its runtime.
\par
The extraction process relies on TASO as the cost model. TASO receives the operator specifications and measures the operator runtime on the underlying hardware. The runtime of an entire computation graph is calculated by summing over all operator costs. In addition to TASO's internal hashing functionality for individual operator configurations, we also store each e-graph with its extracted cost to speed-up the simulation phase.
\par
Once MCTS has exhausted the user-defined search budget, the best rewrite rule is determined based on the root's child node with the highest average value. This rewrite rule is then applied to the main e-graph. Single-pattern rewrite rules are applied by egg, for multi-pattern rewrite rules we rely on TENSAT's efficient search algorithm. The construction phase terminates once the e-graph has saturated or reached the memory limit. Afterwards, the optimized tensor program is extracted from the e-graph. Depending on the size of the e-graph, it is often feasible to use an ILP extractor for this final step.

\section{Evaluation}
\label{sec:experiments}

In this section, we present our experimental results. We begin with an overview of the experimental setup, followed by an analysis of how our cost function affects MCTS optimization performance. Then, we compare our optimizer's performance with TENSAT.

\subsection{Experimental Setup}
\label{sec:experiments_setup}
We run our experiments on 13 models: BERT \cite{bert}, Inception-v3 \cite{inceptionv3}, MobileNet-v2 \cite{mobilenetv2}, NASNet-A \cite{nasnet}, NASRNN \cite{nasrnn}, ResNet-50 \cite{resnet50}, ResNeXt-50 \cite{resnext50}, SqueezeNet \cite{squeezenet}, VGG-19 \cite{vgg}, Transformer-Transducer (TT) \cite{tt}, ViT-Base, ViT-Large, and ViT-Huge \cite{vit}. Similar to \cite{taso, tensat}, we focus our evaluation on model inference as model training requires the storage of intermediate tensors for backpropagation, which generally prevents the graph transformations from being applied directly. Our optimizer supports 30 operators and uses TASO's rewrite rule set comprising of 124 single-pattern and 15 multi-pattern rewrite rules. For NASRNN we had to deactivate one and for TT two multi-pattern rewrite rules, because TASO could not measure the runtime of the resulting computation graph on our hardware. We ran the experiments on an Intel Xeon Silver 4210R CPU @ 2.40GHz with 8 cores and 64 GB RAM and TASO used an NVIDIA A100 80GB GPU to measure the operator runtimes. On a subset of the models, the first 9 listed above, we repeated the experiments on an NVIDIA P100 16GB GPU to evaluate how the hardware impacts the optimization results. For experiments involving an ILP, we used the same solver as TENSAT, SCIP \cite{scip}. 
\par
Due to the stochasticity of the runtime measurements, we repeated all experiments five times and are reporting the mean and standard deviation across all runs. The search budget was set to 128, the maximum simulation depth to 10, and the e-graph construction was stopped once a rewrite rule application resulted in an e-graph of 2,000 or more e-nodes. In the experiments with TENSAT, we found that the default setting of $k_{multi} = 1$ does not always allow TENSAT to reach this node limit. To enable a fair comparison with our approach, we increased $k_{multi}$ for each architecture until the corresponding e-graph either saturated or hit the node limit. 

\subsection{Extraction}
\label{sec:experiments_extraction}

\begin{figure}[]
\includegraphics[width=\columnwidth]{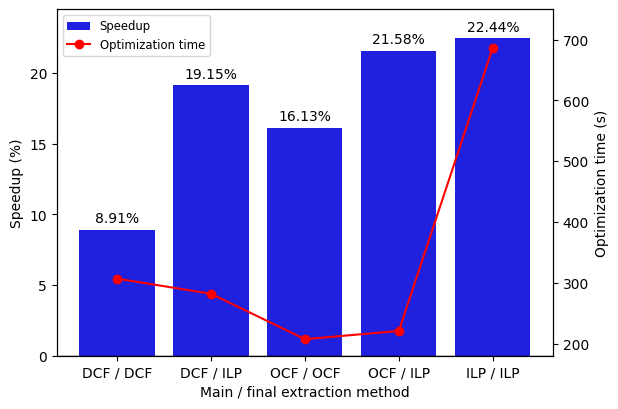}
    \caption{Speedup comparison on an NVIDIA A100 between different main and final extraction methods based on the original and optimized graph runtimes averaged across all runs and models. DCF = default cost function from egg, OCF = our cost function.}
    \label{fig:extraction_speedup}
\vspace{-6pt}
\end{figure}

In Section \ref{sec:methodology_extraction}, we have shown that our cost function is able to significantly improve the runtime estimates for various neural networks. We now analyze how this improved accuracy affects the downstream optimization performance. Figures \ref{fig:extraction_speedup} and \ref{fig:extraction_speedup_p100} compare the runtime speedups and optimization times achieved by MCTS based on different combinations of main and final extraction method on the NVIDIA A100 and P100, respectively. The main method is used during e-graph construction to provide MCTS with a reward signal and the final method is used to extract the output program once e-graph construction has finished. A detailed
breakdown of the results for all architectures can be found in Tables \ref{tab:comprehensive_mcts_results} and \ref{tab:comprehensive_mcts_results_p100}.
\par
Using the default cost function for both the main and final extraction step (DCF/DCF) produces the worst performance. For one model, TT on A100 and NASNet-A on P100, the cost function even increases the original graph runtime by more than 2x. This is consistent with the findings from Yang et al., 2021 \cite{tensat} and confirms that greedy extractors are by default ill-suited to extract tensor programs with shared subgraphs. OCF/OCF evades this failure mode and achieves an average 7-15\% higher speedup. Using an ILP instead of a greedy extractor for the final extraction step further improves the output programs' runtime in both cases (DCF/ILP, OCF/ILP). Even with improved accuracy, greedy extractors remain heuristics that cannot provide the same performance guarantees as ILPs.
\par
Nevertheless, the small difference in obtained speedups between OCF/ILP and ILP/ILP ($<2\%$) shows that greedy extractors using our cost function come close to matching the downstream optimization performance of ILPs while being on average 3-6x faster. Given that the optimization time of ILPs increases exponentially with the size of the e-graph, our proposed cost function introduces greedy extractors as a compelling alternative, especially when working with larger e-graphs. Although the absolute numbers vary between the two hardware backends, the overall findings and the relative ranking of the different extraction methods is consistent.

\subsection{MCTS vs. TENSAT}
\label{sec:experiments_latency}

\begin{figure}[ht]
\includegraphics[width=\columnwidth]{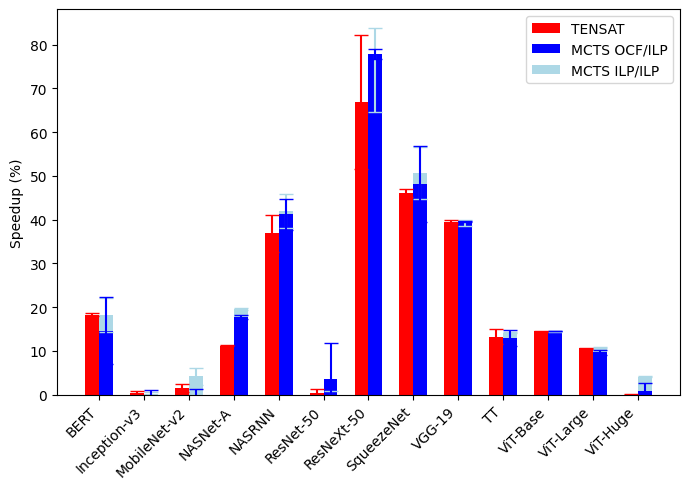}
    \caption{Speedup comparison on an NVIDIA A100 between TENSAT and MCTS based on the original and optimized graph runtimes averaged across five runs}
    \label{fig:tensat_vs_mcts_speedup}
\vspace{-6pt}
\end{figure}

\begin{figure*}[h]
\includegraphics[width=0.95\textwidth]{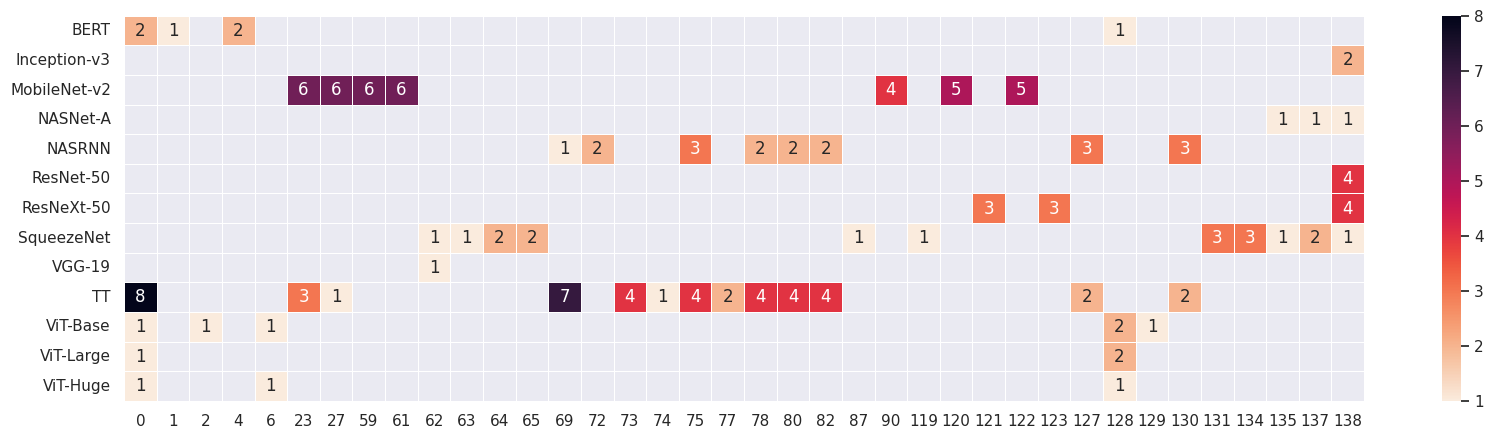}
    \caption{Heatmap showing the number of times TENSAT decided to apply each rewrite rule. 38 out of 139 available rewrite rules were used. TENSAT follows a sequential approach - in each iteration, the multi-pattern rules (124-138) are applied before the single-pattern ones (0-123). Within each set, rules are applied in ascending order.}
    \label{fig:tensat_heatmap}
\end{figure*}

\begin{figure*}[h]
\includegraphics[width=0.95\textwidth]{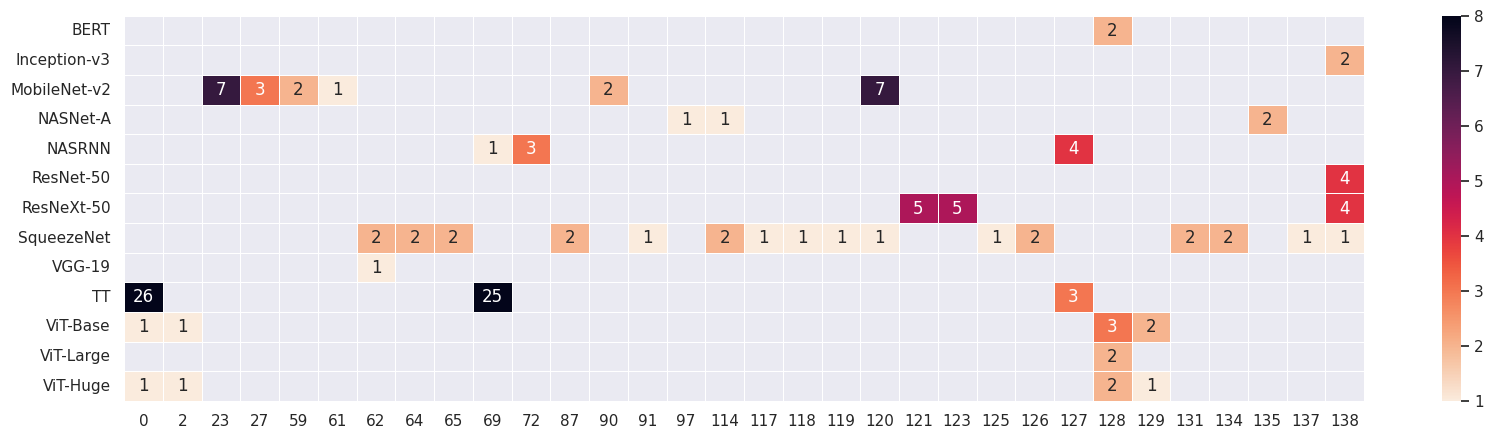}
    \caption{Example heatmap for one run showing the number of times MCTS ILP/ILP decided to apply each rewrite rule. 32 out of 139 available rewrite rules were used.}
    \label{fig:mcts_heatmap}
\end{figure*}

In this section, we compare the end-to-end optimization performances of MCTS and TENSAT. For the purpose of this comparison, we focus on the two best-performing MCTS methods from the previous section - MCTS OCF/ILP and MCTS ILP/ILP. Figure \ref{fig:tensat_vs_mcts_speedup} shows the runtime speedups obtained across the 13 neural network architectures on the NVIDIA A100 with TENSAT as the baseline. More detailed quantitative results, including the optimization times, can be found in Table \ref{tab:comprehensive_mcts_results}. To gain qualitative insights into the decision-making processes of each approach, we plot the rewrite rule applications for one exemplary run in Figures \ref{fig:tensat_heatmap}, \ref{fig:mcts_heatmap}, and \ref{fig:mcts_ocf_ilp_heatmap}.
\par

\par
For three of the models, \textbf{Inception-v3}, \textbf{ResNet-50}, and \textbf{VGG-19} there was only one applicable rewrite rule and thus no possibility for the optimizers to make different decision. Based on TASO's rewrite rule set, the phase ordering problem did not arise in these instances. 
\par
On \textbf{TT}, \textbf{Bert}, \textbf{ViT-Base}, and \textbf{ViT-Large}, MCTS and TENSAT achieve comparable speedups. For the latter three models, the optimizers correctly identify that rewrite rule 128 can significantly reduce the models' runtime. The rewrite rule, visualized in Figure \ref{fig:rule_128}, can eliminate one of two matrix multiplications when specific inputs remain fixed at inference time.
\par
For \textbf{MobileNet-v2}, \textbf{SqueezeNet}, and \textbf{ViT-Huge}, MCTS ILP/ILP is able to achieve 2.5-5\% higher speedups than TENSAT. Comparing the optimizers performance on the three vision transformers (ViT-Base, ViT-Large, ViT-Huge) shows that larger computation graphs - and therefore also larger input e-graphs - benefit MCTS. The reason for this is that the optimizers can apply less rewrite rules before they hit the memory limit and thus the value of every single rewrite rule application increases. TENSAT's sequential selection policy results in the phase-ordering problem whereas MCTS can mitigiate it by identifying the most promising rewrite rules.
\par
On \textbf{NasNet-A}, \textbf{NASRNN}, and \textbf{ResNeXt-50}, MCTS is significantly outperforming TENSAT achieving additional speedups of over 5\%. The biggest improvement was obtained on ResNeXt-50 with an extra speedup of \textasciitilde 11\% followed by NASNet-A with \textasciitilde 8.5\%. These overperformances are achieved by identifying and selecting particularly promising rewrite rules. While TENSAT's sequential selection strategy uses a total of 38 rewrite rules, MCTS ILP/ILP and MCTS OCF/ILP focus on only 32 and 33 rules, respectively, in the runs shown. A good example of the different selection strategies is NASRNN. While TENSAT uses 8 rules in total, both MCTS methods focus on 3 rules that are especially effective at reducing the model's latency and thereby achieve a more than 5\% higher speedup.
\par
These findings are similar to the ones obtained on the NVIDIA P100, which are given in Figure \ref{fig:mcts_vs_tensat_p100} and Table \ref{tab:comprehensive_mcts_results_p100}. The two optimizers achieve comparable performances on 5/9 models and MCTS ILP/ILP is outperforming TENSAT on the other 4 (\textbf{BERT}, \textbf{NASNet-A}, \textbf{NASRNN}, \textbf{SqueezeNet}) by up to 11\%. This highlights the hardware-independent nature of our approach.
\par
MCTS is a way to trade-off between optimization time and performance. If we set the search budget to 1, we recover TENSAT's default behaviour. If we increase the search budget, we obtain a better performance at the cost of longer optimization times. TENSAT does not enable this trade-off, thus MCTS provides a more flexible optimization framework. 

\section{Conclusion and Future Work}
\label{sec:conclusion_and_future_work}

In this paper, we have shown that MCTS can significantly improve the performance of equality saturation-based tensor optimizers by mitigating the phase-ordering problem during e-graph construction. Furthermore, we have devised a novel cost function which enables greedy extractors to take common subexpressions into account and thereby improve the extraction results.
\par
There are several promising avenues for future research to build on our work. Most neural networks consists of a few distinct blocks that are stacked on top of each other. These recurring structures could be exploited by splitting the input graph into its distinct parts, optimizing each part individually before reassembling the optimized graph. This approach could not only shorten the optimization time but also simplify the problem and make it feasible to use learning-based approaches like AlphaZero. In addition to the graph structure, the model latency also depends on the data layouts. Past work \cite{taso} has shown that simultaneously optimizing graph substitutions and data layouts can result in significant speed-ups. We think the incorporation of data layouts into the MCTS paradigm could have a similar effect and result in further runtime reductions.

\clearpage

\bibliographystyle{ACM-Reference-Format}
\bibliography{references}

\clearpage

\appendix

\section{Artifact}

\subsection{Abstract}
Our artifact contains the code for our proposed tensor program optimizer which uses equality saturation and Monte Carlo tree search to reduce the runtime of deep learning models. Our optimizer takes as input a tensor computation graph and rewrite rule set, and outputs an optimized computation graph. The implementation builds on top of several open-source projects: TASO, egg, TENSAT, and MCTS-GEB. Our artifact includes detailed instructions, a Dockerfile to automate large parts of the setup process, the deep learning models to benchmark our approach, and a Jupyter notebook to reproduce the Tables and Figures in our paper. To run, the artifact requires an NVIDIA GPU, Ubuntu 22.04 LTS, NVIDIA drivers, and the NVIDIA Container Toolkit.

\subsection{Artifact check-list (meta-information)}

{\small
\begin{itemize}
  \item {\bf Algorithms: } A tensor program optimizer using equality saturation and Monte Carlo tree search, and a novel cost function for greedy extractors that takes common subexpressions into account.
  \item {\bf Model: } We evaluate and compare our approach on 13 neural network architectures: BERT, Inception-v3, MobileNet-v2, NASNet-A, NASRNN, ResNet-50, ResNeXt-50, SqueezeNet, VGG-19, Transformer-Transducer, ViT-Base, ViT-Large, and ViT-Huge. The models are included in the repository.
  \item {\bf Run-time environment: } Ubuntu 22.04 LTS, NVIDIA driver, NVIDIA Container Toolkit.
  \item {\bf Hardware: } An NVIDIA GPU is required. We used an NVIDIA A100 80GB GPU and an NVIDIA P100 16GB GPU for our experiments.
  \item {\bf Metrics: } Original graph runtime, optimized graph runtime, optimization time, rewrite rule applications.
  \item {\bf Output: } Each experiment outputs, among other things, .txt files with the above-mentioned metrics. We provide a Jupyter notebook to aggregate and process the results.
  \item {\bf Experiments: } We provide a README with a step-by-step installation guide and a Dockerfile to automate large parts of the setup process.
  \item {\bf How much disk space required (approximately)?: } 20GB.
  \item {\bf How much time is needed to prepare workflow (approximately)?: } 1 hour.
  \item {\bf How much time is needed to complete experiments (approximately)?: } 25 hours.
  \item {\bf Publicly available?: } Yes, the artifact is publicly available at \newline https://doi.org/10.5281/zenodo.13278551.
  \item {\bf Code license: } MIT license.
  \item {\bf Archived?: } Yes, the code has been archived at \newline https://doi.org/10.5281/zenodo.13278551.
\end{itemize}
}

\subsection{Description}

\subsubsection{How to access}
The artifact is publicly available at \newline https://doi.org/10.5281/zenodo.13278551.

\subsubsection{Hardware dependencies}
An NVIDIA GPU is required. We used an NVIDIA A100 80GB GPU and an NVIDIA P100 16GB GPU for our experiments.

\subsubsection{Software dependencies}
Ubuntu 22.04 LTS, NVIDIA driver, NVIDIA Container Toolkit.

\subsubsection{Models}
We evaluate and compare our approach on 13 neural network architectures: BERT, Inception-v3, MobileNet-v2, NASNet-A, NASRNN, ResNet-50, ResNeXt-50, SqueezeNet, VGG-19, Transformer-Transducer, ViT-Base, ViT-Large, and ViT-Huge. The models are included in the repository.

\subsection{Installation}
The repository's README provides a step-by-step installation guide. The repository also includes a Dockerfile to automate large parts of the setup process.

\subsection{Experiment workflow}
In our experiments, we evaluate a) our proposed cost function for greedy extractors and b) compare the performance of our tensor program optimizer to TENSAT. We repeat all experiments five times to account for the randomness of MCTS and the stochasticity of the cost model. We provide shell scripts that can automatically run experiments across different cost functions, neural network architectures, and seeds.

\subsection{Evaluation and expected results}
\begin{figure*}[h]
\includegraphics[width=\textwidth]{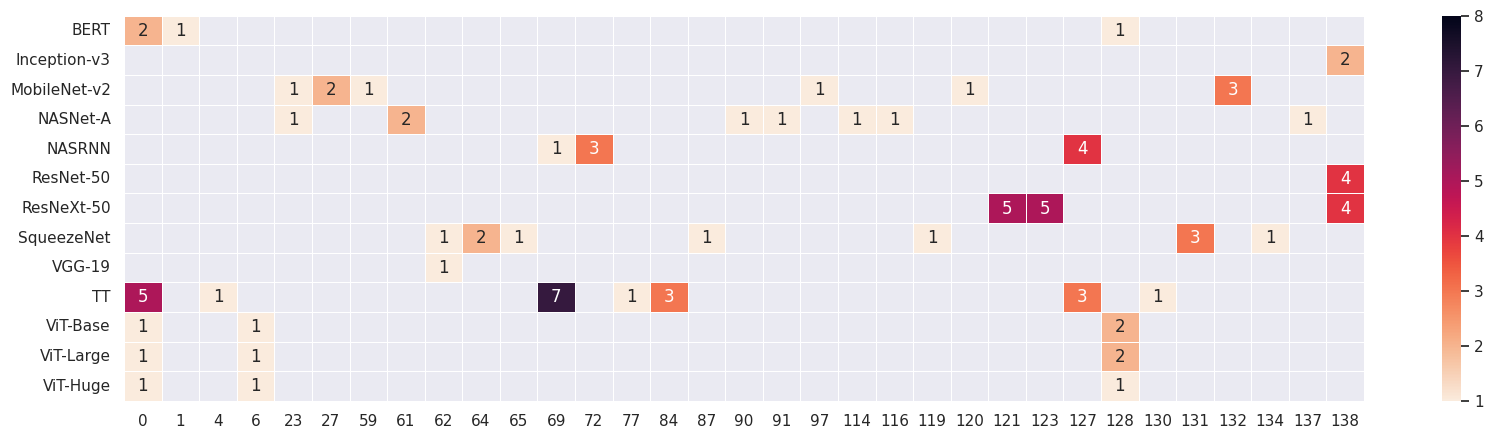}
    \caption{Heatmap showing the number of times MCTS OCF/ILP decided to apply each rewrite rule. 33 out of 139 available rewrite rules were used.}
    \label{fig:mcts_ocf_ilp_heatmap}
\end{figure*}

Each experiment produces the following outputs:
\begin{itemize}
    \item Optimization results including the original graph runtime, optimized graph runtime, and optimization time.
    \item Detailed information on each iteration including which single- and multi-pattern rewrite rules were applied to the e-graph.
    \item Serialized versions of the input and output tensor program.
    \item Visualizations of the original and final e-graph.
\end{itemize}
The results provide insights into the performance of our tensor program optimizer and show the effectiveness of our proposed cost function for greedy extractors. We expect outputs similar to Figures \ref{fig:extraction_speedup}, \ref{fig:tensat_vs_mcts_speedup}, \ref{fig:tensat_heatmap}, \ref{fig:mcts_heatmap}, \ref{fig:mcts_ocf_ilp_heatmap}, and Table \ref{tab:comprehensive_mcts_results}. We provide a Jupyter notebook to reproduce the Tables and Figures.

\subsection{Experiment customization}
There are many ways to customize the experiments. To name just a few: using different rewrite rules sets, modifying TENSAT's $k_{multi}$ parameter, increasing or decreasing the e-graph node limit, changing the MCTS budget, adding neural network architectures, and running the experiments on different GPUs.

\onecolumn
\clearpage
\section{Illustration Cost Function}
\begin{figure*}[h]
    \centering
    \includegraphics[width=0.7\textwidth]{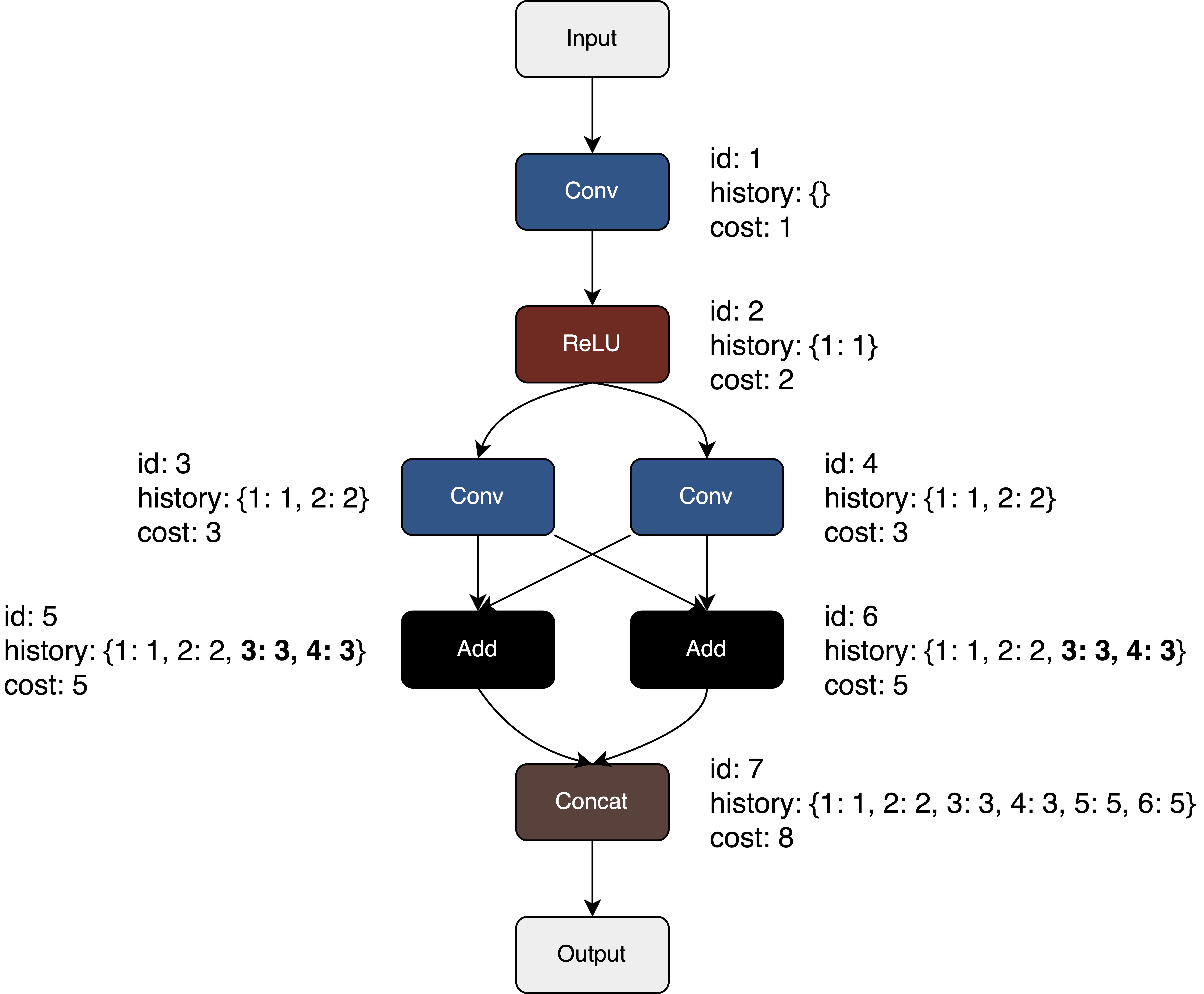}
    \caption{Computation graph for whose initial e-graph our cost function does not produce a correct cost estimate. Assuming a constant operator cost of 1, the correct cost is 7, our cost function returns 8 and existing ones return 15. The inaccuracy is introduced in two steps: 1) The maximum constituent costs of e-classes 5 and 6 is 3 not 4, because their corresponding e-nodes have two children. 2) Therefore, when calculating the cost for the \textit{Concat} node, we assume a common subgraph of cost 3 instead of 4, overestimating the true cost by 1.}
    \label{fig:failure_case}
\end{figure*}

\clearpage
\section{Rewrite Rules}
\subsection{Applications}

\clearpage
\subsection{Examples}

\begin{figure}[h]
    \captionsetup[subfigure]{justification=centering}
     \centering
     \begin{subfigure}[t]{0.45\textwidth}
         \centering
         \includegraphics[width=\textwidth]{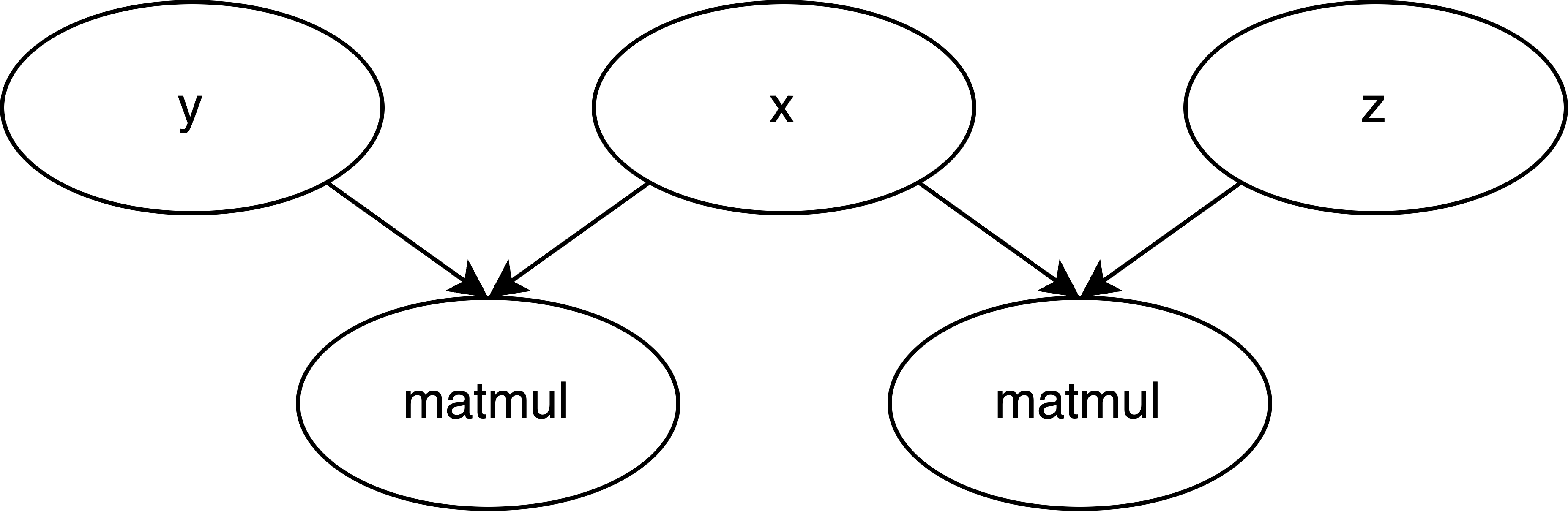}
         \caption{Source pattern}
         \label{fig:rule_128_source_pattern}
     \end{subfigure}
     \hfill
     \begin{subfigure}[t]{0.45\textwidth}
         \centering
         \includegraphics[width=\textwidth]{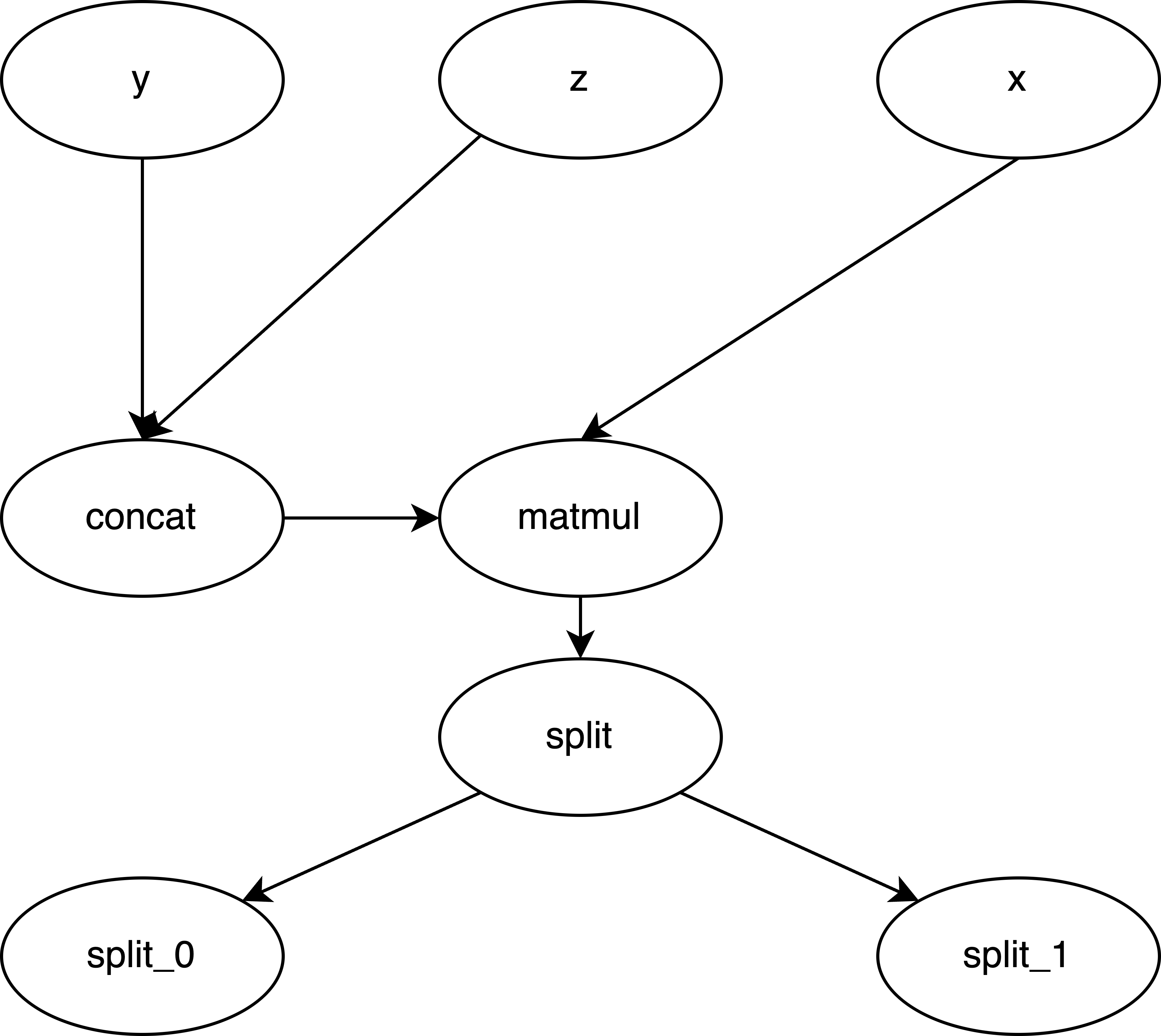}
         \caption{Target pattern}
         \label{fig:rule_128_target_pattern}
     \end{subfigure}
        \caption{Multi-pattern rewrite rule \#128 used in BERT, ViT-Base, ViT-Large, and ViT-Huge. For this visualisation, the two source patterns were merged and constant scalars were omitted. The rule eliminates one of two matrix multiplications. If inputs y and z are fixed (e.g. weights at inference time), they can be preprocessed, which leads to a significant speed-up compared to the source graph.}
        \label{fig:rule_128}
\end{figure}

\begin{figure}[h]
    \captionsetup[subfigure]{justification=centering}
     \centering
     \begin{subfigure}[t]{0.45\textwidth}
         \centering
         \includegraphics[width=\textwidth]{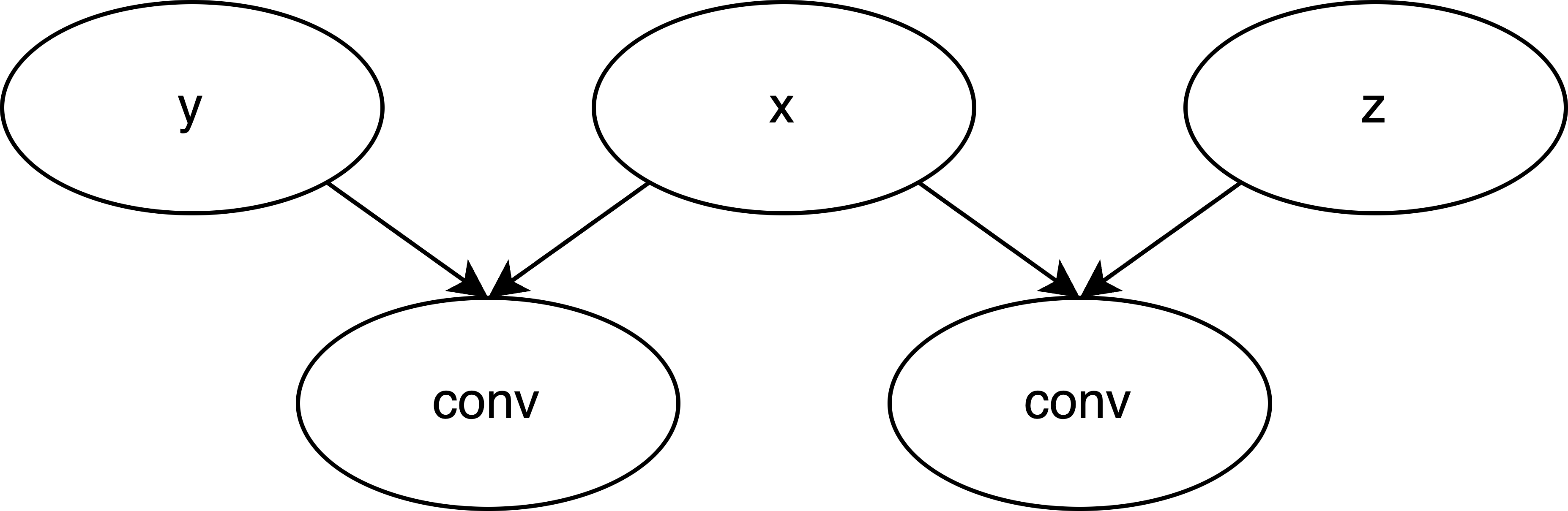}
         \caption{Source pattern}
         \label{fig:rule_138_source_pattern}
     \end{subfigure}
     \hfill
     \begin{subfigure}[t]{0.45\textwidth}
         \centering
         \includegraphics[width=\textwidth]{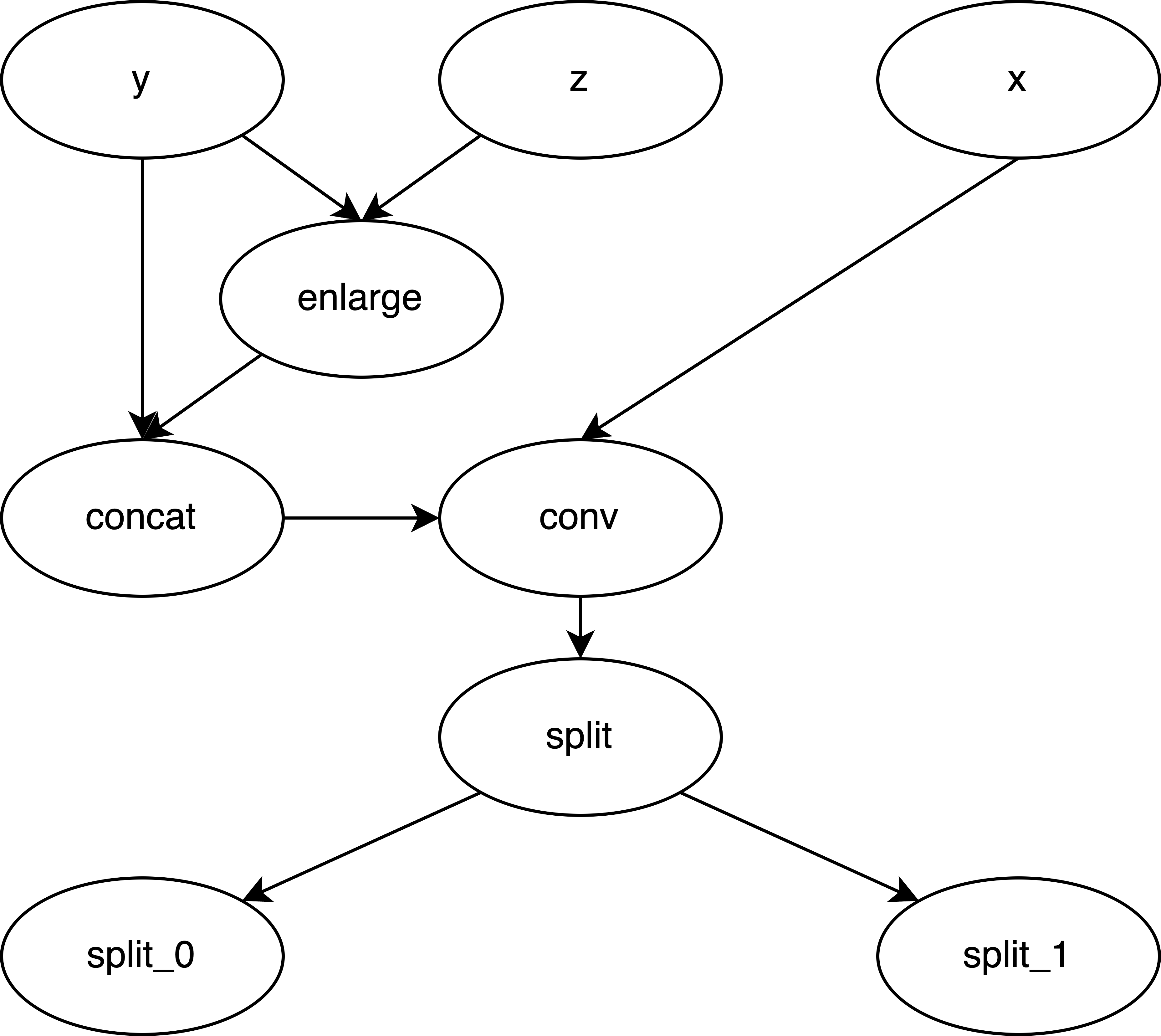}
         \caption{Target pattern}
         \label{fig:rule_138_target_pattern}
     \end{subfigure}
        \caption{Multi-pattern rewrite rule \#138 used in Inception-v3, NASNet-A, ResNet-50, ResNeXt-50, and SqueezeNet. For this visualisation, the two source patterns were merged and constant scalars were omitted. The rule eliminates one of two convolution operations. If inputs y and z are fixed (e.g. weights at inference time), they can be preprocessed, which leads to a significant speed-up compared to the source graph.}
        \label{fig:rule_138}
\end{figure}

\clearpage
\section{Experimental Results}
\input{res/tables/mcts_tensat_comprehensive_a100}
\input{res/tables/mcts_tensat_comprehensive_a100_2}

\clearpage

\noindent
\input{res/tables/extraction_methods_base_cost_comparison_p100}

\vspace{\stretch{1}}

\noindent
\centering
\includegraphics[width=0.6\textwidth]{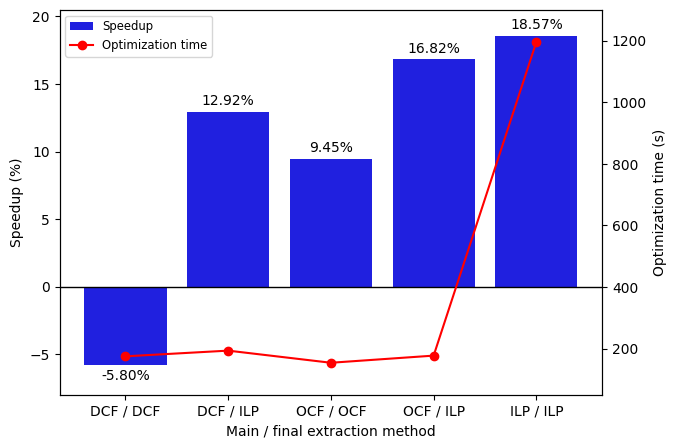}
\captionof{figure}{Speedup comparison on an NVIDIA P100 between
different main and final extraction methods based on the
original and optimized graph runtimes averaged across all
runs and models. DCF = default cost function from egg, OCF
= our cost function.}
\label{fig:extraction_speedup_p100}

\vspace{\stretch{1}} %

\noindent
\centering
\includegraphics[width=0.6\textwidth]{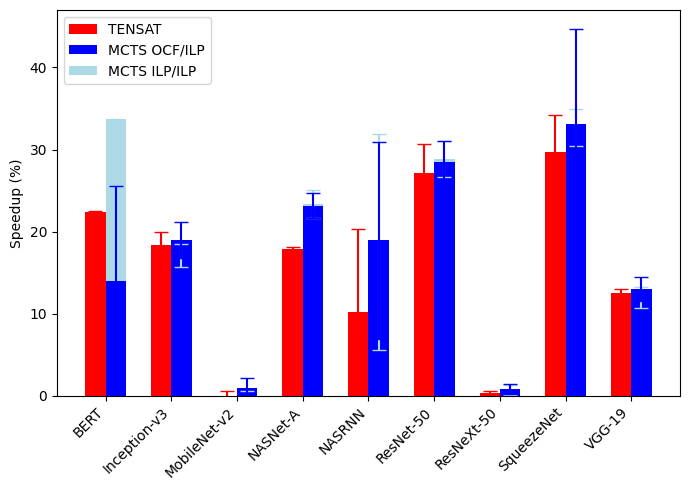}
\captionof{figure}{Speedup comparison on an NVIDIA P100 between TENSAT and MCTS based on the original and optimized graph runtimes averaged across five runs.}
\label{fig:mcts_vs_tensat_p100}

\input{res/tables/mcts_tensat_comprehensive_p100}

\end{document}

%% file: res/tables/extraction_methods_base_cost_comparison_a100.tex
\begin{table*}[h]
\centering
\caption{Predicted runtimes on an NVIDIA A100 by different extraction methods based on the initial e-graph of 13 models. Except for VGG-19, a greedy extractor using a default cost function significantly overestimates the initial graph runtime. Using our cost function, the greedy extractor matches the accuracy of an ILP extractor on all architectures except NasNet-A.}
\label{tab:comparison_extraction_methods}
\resizebox{\textwidth}{!}{%
\begin{tabular}{@{}llllllllllllll@{}}
\toprule
Architecture          & BERT & Inception-v3 & MobileNet-v2 & NasNet-A              & NASRNN  & ResNet-50 & ResNeXt-50 & SqueezeNet & VGG-19 & TT         & ViT-Base   & ViT-Large             & ViT-Huge              \\ \midrule
ILP                   & 0.93 & 0.79         & 1.06         & 3.45                  & 2.48    & 0.29      & 1.26       & 0.57       & 0.61   & 3.25       & 1.93       & 5.15                  & 11.65                 \\
Default cost function & 2.75 & 410938.5     & 421.32       & $6.85 \times 10^{12}$ & 2319.06 & 1565.76   & 15162.62   & 41.51      & 0.61   & 6311745500 & 2982248400 & $2.55 \times 10^{20}$ & $6.44 \times 10^{27}$ \\
Our cost function     & 0.93 & 0.79         & 1.06         & 3.59                  & 2.48    & 0.29      & 1.26       & 0.57       & 0.61   & 3.25       & 1.93       & 5.15                  & 11.65                 \\ \bottomrule
\end{tabular}%
}
\end{table*}

%% file: res/tables/mcts_tensat_comprehensive_a100.tex
\begin{table*}[h]
\centering
\caption{Comparison of MCTS with different extraction methods and TENSAT as baseline on an NVIDIA A100. The default value for $k_{multi}$ of 1 was increased until TENSAT's e-graph saturated or hit the node limit. The original graph runtimes may vary due to the stochasticity of the cost model and are given for reference. Runtime speedups are calculated based on the original and optimized graph runtimes. Default greedy = greedy extractor with default cost function, our greedy = greedy extractor with our cost function.}
\label{tab:comprehensive_mcts_results}
\resizebox{\textwidth}{!}{%
\begin{tabular}{@{}lllllllll@{}}
\toprule
Model &
  Approach &
  $k_{multi}$ &
  Main extraction &
  Final extraction &
  Original runtime (ms) &
  Optimized runtime (ms) &
  Runtime speedup (\%) &
  Optimization time (s) \\ \midrule
\multirow{6}{*}{BERT} &
  \multirow{5}{*}{MCTS} &
  - &
  Default greedy &
  Default greedy &
  0.92 $\pm$ 0.00 &
  0.93 $\pm$ 0.00 &
  -0.16 $\pm$ 0.16 &
  43.00 $\pm$ 4.47 \\
 &
   &
  - &
  Default greedy &
  ILP &
  0.93 $\pm$ 0.01 &
  0.92 $\pm$ 0.01 &
  0.72 $\pm$ 1.55 &
  41.60 $\pm$ 1.52 \\
 &
   &
  - &
  Our greedy &
  Our greedy &
  0.93 $\pm$ 0.00 &
  0.93 $\pm$ 0.01 &
  -0.10 $\pm$ 0.22 &
  46.60 $\pm$ 5.81 \\
 &
   &
  - &
  Our greedy &
  ILP &
  0.93 $\pm$ 0.00 &
  0.79 $\pm$ 0.07 &
  14.63 $\pm$ 7.65 &
  43.80 $\pm$ 7.60 \\
 &
   &
  - &
  ILP &
  ILP &
  0.93 $\pm$ 0.00 &
  0.76 $\pm$ 0.04 &
  18.20 $\pm$ 3.92 &
  1178.60 $\pm$ 395.89 \\
 &
  TENSAT &
  1 &
  - &
  ILP &
  0.93 $\pm$ 0.00 &
  0.76 $\pm$ 0.00 &
  \textbf{18.27 $\pm$ 0.33} &
  \textbf{1.14 $\pm$ 0.06} \\ \midrule
\multirow{6}{*}{Inception-v3} &
  \multirow{5}{*}{MCTS} &
  - &
  Default greedy &
  Default greedy &
  0.36 $\pm$ 0.00 &
  0.36 $\pm$ 0.00 &
  -0.05 $\pm$ 0.30 &
  3.00 $\pm$ 0.00 \\
 &
   &
  - &
  Default greedy &
  ILP &
  0.36 $\pm$ 0.00 &
  0.36 $\pm$ 0.00 &
  0.51 $\pm$ 0.47 &
  3.00 $\pm$ 0.00 \\
 &
   &
  - &
  Our greedy &
  Our greedy &
  0.37 $\pm$ 0.01 &
  0.36 $\pm$ 0.00 &
  \textbf{1.06 $\pm$ 1.68} &
  3.00 $\pm$ 0.00 \\
 &
   &
  - &
  Our greedy &
  ILP &
  0.36 $\pm$ 0.00 &
  0.36 $\pm$ 0.00 &
  -0.01 $\pm$ 1.15 &
  3.40 $\pm$ 0.55 \\
 &
   &
  - &
  ILP &
  ILP &
  0.36 $\pm$ 0.00 &
  0.36 $\pm$ 0.00 &
  0.25 $\pm$ 0.26 &
  3.00 $\pm$ 0.00 \\
 &
  TENSAT &
  2 &
  - &
  ILP &
  0.36 $\pm$ 0.00 &
  0.36 $\pm$ 0.00 &
  0.36 $\pm$ 0.56 &
  \textbf{0.89 $\pm$ 0.07} \\ \midrule
\multirow{6}{*}{MobileNet-v2} &
  \multirow{5}{*}{MCTS} &
  - &
  Default greedy &
  Default greedy &
  0.85 $\pm$ 0.03 &
  0.82 $\pm$ 0.01 &
  4.05 $\pm$ 2.55 &
  1149.80 $\pm$ 144.60 \\
 &
   &
  - &
  Default greedy &
  ILP &
  0.87 $\pm$ 0.06 &
  0.84 $\pm$ 0.08 &
  2.94 $\pm$ 11.30 &
  992.00 $\pm$ 328.22 \\
 &
   &
  - &
  Our greedy &
  Our greedy &
  0.85 $\pm$ 0.01 &
  0.88 $\pm$ 0.01 &
  -3.61 $\pm$ 1.06 &
  419.40 $\pm$ 219.31 \\
 &
   &
  - &
  Our greedy &
  ILP &
  0.84 $\pm$ 0.01 &
  0.85 $\pm$ 0.01 &
  -0.36 $\pm$ 1.60 &
  387.60 $\pm$ 108.86 \\
 &
   &
  - &
  ILP &
  ILP &
  0.84 $\pm$ 0.01 &
  0.81 $\pm$ 0.01 &
  \textbf{4.16 $\pm$ 2.03} &
  754.00 $\pm$ 225.11 \\
 &
  TENSAT &
  1 &
  - &
  ILP &
  0.84 $\pm$ 0.01 &
  0.83 $\pm$ 0.01 &
  1.52 $\pm$ 1.00 &
  \textbf{0.93 $\pm$ 0.03} \\ \midrule
\multirow{6}{*}{NASNet-A} &
  \multirow{5}{*}{MCTS} &
  - &
  Default greedy &
  Default greedy &
  2.87 $\pm$ 0.03 &
  2.35 $\pm$ 0.14 &
  18.29 $\pm$ 4.07 &
  898.00 $\pm$ 185.95 \\
 &
   &
  - &
  Default greedy &
  ILP &
  2.89 $\pm$ 0.00 &
  2.27 $\pm$ 0.10 &
  \textbf{21.46 $\pm$ 3.55} &
  777.60 $\pm$ 107.27 \\
 &
   &
  - &
  Our greedy &
  Our greedy &
  2.87 $\pm$ 0.03 &
  2.39 $\pm$ 0.02 &
  16.55 $\pm$ 0.18 &
  1043.60 $\pm$ 60.66 \\
 &
   &
  - &
  Our greedy &
  ILP &
  2.85 $\pm$ 0.03 &
  2.35 $\pm$ 0.03 &
  17.71 $\pm$ 0.50 &
  1114.40 $\pm$ 67.12 \\
 &
   &
  - &
  ILP &
  ILP &
  2.86 $\pm$ 0.02 &
  2.30 $\pm$ 0.02 &
  19.62 $\pm$ 0.29 &
  305.00 $\pm$ 5.43 \\
 &
  TENSAT &
  1 &
  - &
  ILP &
  2.87 $\pm$ 0.03 &
  2.55 $\pm$ 0.02 &
  11.18 $\pm$ 0.08 &
  \textbf{0.45 $\pm$ 0.03} \\ \midrule
\multirow{6}{*}{NASRNN} &
  \multirow{5}{*}{MCTS} &
  - &
  Default greedy &
  Default greedy &
  2.50 $\pm$ 0.16 &
  2.25 $\pm$ 0.03 &
  9.75 $\pm$ 5.49 &
  102.20 $\pm$ 4.09 \\
 &
   &
  - &
  Default greedy &
  ILP &
  2.50 $\pm$ 0.07 &
  1.50 $\pm$ 0.02 &
  39.96 $\pm$ 1.22 &
  107.60 $\pm$ 6.99 \\
 &
   &
  - &
  Our greedy &
  Our greedy &
  2.49 $\pm$ 0.06 &
  1.98 $\pm$ 0.45 &
  20.65 $\pm$ 17.77 &
  200.80 $\pm$ 156.27 \\
 &
   &
  - &
  Our greedy &
  ILP &
  2.58 $\pm$ 0.15 &
  1.51 $\pm$ 0.04 &
  41.19 $\pm$ 3.50 &
  114.60 $\pm$ 2.61 \\
 &
   &
  - &
  ILP &
  ILP &
  2.63 $\pm$ 0.21 &
  1.52 $\pm$ 0.05 &
  \textbf{42.07 $\pm$ 3.91} &
  2782.00 $\pm$ 1719.55 \\
 &
  TENSAT &
  3 &
  - &
  ILP &
  2.47 $\pm$ 0.04 &
  1.56 $\pm$ 0.10 &
  36.92 $\pm$ 4.24 &
  \textbf{9.36 $\pm$ 0.74} \\ \midrule
\multirow{6}{*}{ResNet-50} &
  \multirow{5}{*}{MCTS} &
  - &
  Default greedy &
  Default greedy &
  0.26 $\pm$ 0.00 &
  0.26 $\pm$ 0.00 &
  -0.23 $\pm$ 0.26 &
  9.60 $\pm$ 0.55 \\
 &
   &
  - &
  Default greedy &
  ILP &
  0.28 $\pm$ 0.03 &
  0.28 $\pm$ 0.03 &
  -1.24 $\pm$ 17.80 &
  9.80 $\pm$ 0.45 \\
 &
   &
  - &
  Our greedy &
  Our greedy &
  0.27 $\pm$ 0.02 &
  0.26 $\pm$ 0.00 &
  2.52 $\pm$ 6.21 &
  10.00 $\pm$ 0.00 \\
 &
   &
  - &
  Our greedy &
  ILP &
  0.28 $\pm$ 0.03 &
  0.26 $\pm$ 0.00 &
  \textbf{3.60 $\pm$ 8.25} &
  10.00 $\pm$ 0.00 \\
 &
   &
  - &
  ILP &
  ILP &
  0.26 $\pm$ 0.00 &
  0.26 $\pm$ 0.00 &
  -0.22 $\pm$ 0.97 &
  23.20 $\pm$ 7.79 \\
 &
  TENSAT &
  4 &
  - &
  ILP &
  0.26 $\pm$ 0.00 &
  0.26 $\pm$ 0.00 &
  0.43 $\pm$ 0.94 &
  \textbf{2.96 $\pm$ 0.66} \\ \midrule
\multirow{6}{*}{ResNeXt-50} &
  \multirow{5}{*}{MCTS} &
  - &
  Default greedy &
  Default greedy &
  1.05 $\pm$ 0.24 &
  0.27 $\pm$ 0.00 &
  72.69 $\pm$ 9.40 &
  38.20 $\pm$ 4.97 \\
 &
   &
  - &
  Default greedy &
  ILP &
  1.11 $\pm$ 0.31 &
  0.27 $\pm$ 0.00 &
  72.99 $\pm$ 11.90 &
  38.80 $\pm$ 4.97 \\
 &
   &
  - &
  Our greedy &
  Our greedy &
  1.24 $\pm$ 0.02 &
  0.27 $\pm$ 0.00 &
  \textbf{78.06 $\pm$ 0.19} &
  97.60 $\pm$ 5.68 \\
 &
   &
  - &
  Our greedy &
  ILP &
  1.26 $\pm$ 0.07 &
  0.28 $\pm$ 0.02 &
  77.84 $\pm$ 1.09 &
  101.40 $\pm$ 5.59 \\
 &
   &
  - &
  ILP &
  ILP &
  1.13 $\pm$ 0.29 &
  0.27 $\pm$ 0.00 &
  74.20 $\pm$ 9.54 &
  138.60 $\pm$ 5.77 \\
 &
  TENSAT &
  4 &
  - &
  ILP &
  1.13 $\pm$ 0.32 &
  0.34 $\pm$ 0.01 &
  66.87 $\pm$ 15.36 &
  \textbf{2.69 $\pm$ 1.08} \\ \midrule
\multirow{6}{*}{SqueezeNet} &
  \multirow{5}{*}{MCTS} &
  - &
  Default greedy &
  Default greedy &
  0.30 $\pm$ 0.01 &
  0.13 $\pm$ 0.00 &
  \textbf{56.41 $\pm$ 0.95} &
  283.40 $\pm$ 254.00 \\
 &
   &
  - &
  Default greedy &
  ILP &
  0.31 $\pm$ 0.01 &
  0.16 $\pm$ 0.02 &
  47.39 $\pm$ 7.05 &
  108.40 $\pm$ 10.26 \\
 &
   &
  - &
  Our greedy &
  Our greedy &
  0.31 $\pm$ 0.01 &
  0.16 $\pm$ 0.02 &
  47.23 $\pm$ 6.60 &
  260.80 $\pm$ 203.44 \\
 &
   &
  - &
  Our greedy &
  ILP &
  0.33 $\pm$ 0.05 &
  0.17 $\pm$ 0.01 &
  48.16 $\pm$ 8.77 &
  337.00 $\pm$ 301.81 \\
 &
   &
  - &
  ILP &
  ILP &
  0.32 $\pm$ 0.03 &
  0.16 $\pm$ 0.02 &
  50.66 $\pm$ 6.01 &
  1450.80 $\pm$ 1437.77 \\
 &
  TENSAT &
  3 &
  - &
  ILP &
  0.30 $\pm$ 0.00 &
  0.16 $\pm$ 0.00 &
  46.02 $\pm$ 0.91 &
  \textbf{1.55 $\pm$ 0.31} \\ \midrule
\multirow{6}{*}{VGG-19} &
  \multirow{5}{*}{MCTS} &
  - &
  Default greedy &
  Default greedy &
  0.64 $\pm$ 0.00 &
  0.39 $\pm$ 0.00 &
  \textbf{39.72 $\pm$ 0.14} &
  1.00 $\pm$ 0.00 \\
 &
   &
  - &
  Default greedy &
  ILP &
  0.64 $\pm$ 0.00 &
  0.39 $\pm$ 0.00 &
  39.63 $\pm$ 0.11 &
  1.00 $\pm$ 0.00 \\
 &
   &
  - &
  Our greedy &
  Our greedy &
  0.64 $\pm$ 0.00 &
  0.39 $\pm$ 0.00 &
  39.69 $\pm$ 0.35 &
  1.00 $\pm$ 0.00 \\
 &
   &
  - &
  Our greedy &
  ILP &
  0.64 $\pm$ 0.00 &
  0.39 $\pm$ 0.00 &
  39.69 $\pm$ 0.12 &
  1.00 $\pm$ 0.00 \\
 &
   &
  - &
  ILP &
  ILP &
  0.64 $\pm$ 0.00 &
  0.39 $\pm$ 0.00 &
  39.21 $\pm$ 0.73 &
  1.00 $\pm$ 0.00 \\
 &
  TENSAT &
  1 &
  - &
  ILP &
  0.64 $\pm$ 0.00 &
  0.39 $\pm$ 0.00 &
  39.58 $\pm$ 0.34 &
  \textbf{0.06 $\pm$ 0.00} \\ \bottomrule
\end{tabular}%
}
\end{table*}

%% file: res/tables/mcts_tensat_comprehensive_a100_2.tex
\begin{table}[]
\centering
\caption*{Table 2 continued from previous page}
\label{tab:comprehensive_mcts_results_2}
\resizebox{\textwidth}{!}{%
\begin{tabular}{@{}lllllllll@{}}
\toprule
Model &
  Approach &
  $k_{multi}$ &
  Main extraction &
  Final extraction &
  Original runtime (ms) &
  Optimized runtime (ms) &
  Runtime speedup (\%) &
  Optimization time (s) \\ \midrule
\multirow{6}{*}{TT} &
  \multirow{5}{*}{MCTS} &
  - &
  Default greedy &
  Default greedy &
  3.11 $\pm$ 0.01 &
  6.95 $\pm$ 0.19 &
  -123.81 $\pm$ 5.35 &
  1202.33 $\pm$ 39.12 \\
 &        & - & Default greedy & ILP        & 3.11 $\pm$ 0.01  & 2.81 $\pm$ 0.16  & 9.68 $\pm$ 4.93           & 1363.00 $\pm$ 8.29       \\
 &        & - & Our greedy     & Our greedy & 3.10 $\pm$ 0.01  & 2.85 $\pm$ 0.03  & 8.06 $\pm$ 1.21           & 436.20 $\pm$ 105.31      \\
 &        & - & Our greedy     & ILP        & 3.11 $\pm$ 0.01  & 2.71 $\pm$ 0.05  & 13.02 $\pm$ 1.78          & 577.00 $\pm$ 290.64      \\
 &        & - & ILP            & ILP        & 3.09 $\pm$ 0.01  & 2.65 $\pm$ 0.01  & \textbf{14.18 $\pm$ 0.19} & 2080.80 $\pm$ 793.04     \\
 & TENSAT & 2 & -              & ILP        & 3.10 $\pm$ 0.01  & 2.68 $\pm$ 0.06  & 13.29 $\pm$ 1.67          & \textbf{1.35 $\pm$ 0.08} \\ \midrule
\multirow{6}{*}{ViT-Base} &
  \multirow{5}{*}{MCTS} &
  - &
  Default greedy &
  Default greedy &
  2.13 $\pm$ 0.01 &
  2.08 $\pm$ 0.01 &
  2.20 $\pm$ 0.21 &
  110.60 $\pm$ 22.57 \\
 &        & - & Default greedy & ILP        & 2.14 $\pm$ 0.01  & 1.96 $\pm$ 0.06  & 8.34 $\pm$ 2.85           & 115.00 $\pm$ 22.00       \\
 &        & - & Our greedy     & Our greedy & 2.13 $\pm$ 0.01  & 2.14 $\pm$ 0.01  & -0.12 $\pm$ 0.20          & 60.60 $\pm$ 8.62         \\
 &        & - & Our greedy     & ILP        & 2.14 $\pm$ 0.00  & 1.83 $\pm$ 0.00  & \textbf{14.60 $\pm$ 0.05} & 61.00 $\pm$ 13.51        \\
 &        & - & ILP            & ILP        & 2.14 $\pm$ 0.01  & 1.83 $\pm$ 0.01  & 14.48 $\pm$ 0.08          & 97.80 $\pm$ 2.77         \\
 & TENSAT & 2 & -              & ILP        & 2.13 $\pm$ 0.01  & 1.83 $\pm$ 0.01  & 14.47 $\pm$ 0.12          & \textbf{0.61 $\pm$ 0.02} \\ \midrule
\multirow{6}{*}{ViT-Large} &
  \multirow{5}{*}{MCTS} &
  - &
  Default greedy &
  Default greedy &
  5.63 $\pm$ 0.01 &
  4.77 $\pm$ 0.01 &
  \textbf{15.24 $\pm$ 0.02} &
  53.40 $\pm$ 6.07 \\
 &        & - & Default greedy & ILP        & 5.62 $\pm$ 0.01  & 5.25 $\pm$ 0.01  & 6.60 $\pm$ 0.03           & 54.20 $\pm$ 6.06         \\
 &        & - & Our greedy     & Our greedy & 5.62 $\pm$ 0.02  & 5.64 $\pm$ 0.02  & -0.20 $\pm$ 0.11          & 52.20 $\pm$ 7.69         \\
 &        & - & Our greedy     & ILP        & 5.63 $\pm$ 0.01  & 5.09 $\pm$ 0.04  & 9.69 $\pm$ 0.58           & 56.40 $\pm$ 1.67         \\
 &        & - & ILP            & ILP        & 5.64 $\pm$ 0.02  & 5.03 $\pm$ 0.01  & 10.77 $\pm$ 0.15          & 35.00 $\pm$ 3.74         \\
 & TENSAT & 2 & -              & ILP        & 5.62 $\pm$ 0.02  & 5.03 $\pm$ 0.01  & 10.61 $\pm$ 0.09          & \textbf{0.69 $\pm$ 0.03} \\ \midrule
\multirow{6}{*}{ViT-Huge} &
  \multirow{5}{*}{MCTS} &
  - &
  Default greedy &
  Default greedy &
  12.12 $\pm$ 0.03 &
  8.67 $\pm$ 0.03 &
  \textbf{28.48 $\pm$ 0.07} &
  60.80 $\pm$ 12.34 \\
 &        & - & Default greedy & ILP        & 12.12 $\pm$ 0.02 & 12.12 $\pm$ 0.01 & 0.02 $\pm$ 0.04           & 69.80 $\pm$ 4.87         \\
 &        & - & Our greedy     & Our greedy & 12.12 $\pm$ 0.03 & 12.13 $\pm$ 0.02 & -0.06 $\pm$ 0.03          & 62.20 $\pm$ 20.62        \\
 &        & - & Our greedy     & ILP        & 12.12 $\pm$ 0.02 & 12.02 $\pm$ 0.23 & 0.82 $\pm$ 1.83           & 62.80 $\pm$ 17.08        \\
 &        & - & ILP            & ILP        & 12.14 $\pm$ 0.02 & 11.64 $\pm$ 0.01 & 4.14 $\pm$ 0.07           & 76.60 $\pm$ 1.95         \\
 & TENSAT & 1 & -              & ILP        & 12.13 $\pm$ 0.02 & 12.13 $\pm$ 0.02 & 0.02 $\pm$ 0.05           & \textbf{0.49 $\pm$ 0.00} \\ \bottomrule
\end{tabular}%
}
\end{table}

%% file: res/tables/extraction_methods_base_cost_comparison_p100.tex
\begin{table*}[t]
\centering
\captionsetup{skip=0.1in}
\caption{Predicted runtimes on an NVIDIA P100 by different extraction methods based on the initial e-graph of 9 models. Except for VGG-19, a greedy extractor using a default cost function significantly overestimates the initial graph runtime. Using our cost function, the greedy extractor matches the accuracy of an ILP extractor on all architectures except NasNet-A.}
\label{tab:comparison_extraction_methods_p100}
\resizebox{\textwidth}{!}{%
\begin{tabular}{@{}llllllllll@{}}
\toprule
Architecture          & BERT & Inception-v3 & MobileNet-v2 & NasNet-A       & NASRNN  & ResNet-50 & ResNeXt-50 & SqueezeNet & VGG-19 \\ \midrule
ILP               & 1.44 & 2.41 & 3.49 & 25.77 & 2.6 & 9.3 & 13.44 & 2.07 & 6.73 \\
Default cost function & 4.93 & 2253.37      & 720.25       & $4.31 \times 10^{13}$ & 2437.84 & 22029.49  & 54887.95   & 125.0      & 6.73   \\
Our cost function & 1.44 & 2.41 & 3.49 & 33.86 & 2.6 & 9.3 & 13.44 & 2.07 & 6.73 \\ \bottomrule
\end{tabular}%
}
\end{table*}

%% file: res/tables/mcts_tensat_comprehensive_p100.tex
\begin{table}[h]
\centering
\caption{Comparison of MCTS with different extraction methods together with TENSAT as baseline on an NVIDIA P100. The default value for $k_{multi}$ of 1 was increased until TENSAT's e-graph either saturated or hit the node limit. The original graph runtimes may vary due to the stochasticity of the cost model and are given for reference. Runtime speedups are calculated based on the original and optimized graph runtimes. Default greedy = greedy extractor with default cost function, our greedy = greedy extractor with our cost function.}
\label{tab:comprehensive_mcts_results_p100}
\resizebox{\textwidth}{!}{%
\begin{tabular}{lllllllll}
\hline
Model &
  Approach &
  $k_{multi}$ &
  Main extraction &
  Final extraction &
  Original runtime (ms) &
  Optimized runtime (ms) &
  Runtime speedup (\%) &
  Optimization time (s) \\ \hline
\multirow{6}{*}{BERT} &
  \multirow{5}{*}{MCTS} &
  - &
  Default greedy &
  Default greedy &
  1.32 $\pm$ 0.0 &
  1.32 $\pm$ 0.0 &
  0.1 $\pm$ 0.1 &
  53.0 $\pm$ 3.54 \\
 &        & - & Default greedy & ILP        & 1.32 $\pm$ 0.0   & 1.30 $\pm$ 0.0   & 1.49 $\pm$ 0.03            & 52.6 $\pm$ 2.07           \\
 &        & - & Our greedy     & Our greedy & 1.32 $\pm$ 0.0   & 1.32 $\pm$ 0.0   & 0.02 $\pm$ 0.07            & 60.8 $\pm$ 6.76           \\
 &        & - & Our greedy     & ILP        & 1.32 $\pm$ 0.0   & 1.13 $\pm$ 0.15  & 14.03 $\pm$ 11.47          & 60.6 $\pm$ 4.28           \\
 &        & - & ILP            & ILP        & 1.32 $\pm$ 0.0   & 0.87 $\pm$ 0.0   & \textbf{33.71 $\pm$ 0.07}  & 1170.6 $\pm$ 1061.79      \\
 & TENSAT & 1 & -              & ILP        & 1.32 $\pm$ 0.0   & 1.03 $\pm$ 0.0     & 22.35 $\pm$ 0.17           & \textbf{2.05 $\pm$ 0.47}  \\ \hline
\multirow{6}{*}{Inception-v3} &
  \multirow{5}{*}{MCTS} &
  - &
  Default greedy &
  Default greedy &
  15.58 $\pm$ 1.14 &
  15.29 $\pm$ 0.79 &
  1.8 $\pm$ 2.1 &
  6.2 $\pm$ 0.45 \\
 &        & - & Default greedy & ILP        & 15.21 $\pm$ 0.61 & 12.78 $\pm$ 0.42 & 15.93 $\pm$ 1.64           & 6.2 $\pm$ 0.45            \\
 &        & - & Our greedy     & Our greedy & 15.18 $\pm$ 0.41 & 15.21 $\pm$ 0.5  & -0.24 $\pm$ 1.02           & 6.6 $\pm$ 0.55            \\
 &        & - & Our greedy     & ILP        & 15.22 $\pm$ 0.36 & 12.34 $\pm$ 0.52 & \textbf{18.94 $\pm$ 2.18}  & 6.2 $\pm$ 0.45            \\
 &        & - & ILP            & ILP        & 15.08 $\pm$ 0.26 & 12.51 $\pm$ 0.29 & 17.06 $\pm$ 1.43           & 7.4 $\pm$ 0.55            \\
 & TENSAT & 2 & -              & ILP        & 15.41 $\pm$ 0.48 & 12.58 $\pm$ 0.25 & 18.34 $\pm$ 1.57           & \textbf{2.84 $\pm$ 0.08}  \\ \hline
\multirow{6}{*}{MobileNet-v2} &
  \multirow{5}{*}{MCTS} &
  - &
  Default greedy &
  Default greedy &
  3.61 $\pm$ 0.01 &
  3.64 $\pm$ 0.02 &
  -0.91 $\pm$ 0.49 &
  573.2 $\pm$ 41.69 \\
 &        & - & Default greedy & ILP        & 3.6 $\pm$ 0.05   & 3.62 $\pm$ 0.05  & -0.64 $\pm$ 0.79           & 596.8 $\pm$ 40.18         \\
 &        & - & Our greedy     & Our greedy & 3.57 $\pm$ 0.02  & 3.58 $\pm$ 0.04  & -0.28 $\pm$ 0.53           & 285.0 $\pm$ 114.4         \\
 &        & - & Our greedy     & ILP        & 3.58 $\pm$ 0.05  & 3.54 $\pm$ 0.05  & \textbf{0.96 $\pm$ 1.18}   & 357.8 $\pm$ 109.9         \\
 &        & - & ILP            & ILP        & 3.58 $\pm$ 0.11  & 3.58 $\pm$ 0.1   & 0.12 $\pm$ 0.46            & 2106.2 $\pm$ 294.56       \\
 & TENSAT & 1 & -              & ILP        & 3.55 $\pm$ 0.06  & 3.58 $\pm$ 0.07  & -0.68 $\pm$ 1.2            & \textbf{3.4 $\pm$ 0.14}   \\ \hline
\multirow{6}{*}{NASNet-A} &
  \multirow{5}{*}{MCTS} &
  - &
  Default greedy &
  Default greedy &
  25.48 $\pm$ 0.12 &
  60.36 $\pm$ 2.85 &
  -136.91 $\pm$ 10.88 &
  487.4 $\pm$ 120.42 \\
 &        & - & Default greedy & ILP        & 25.73 $\pm$ 0.08 & 19.93 $\pm$ 0.45 & 22.53 $\pm$ 1.87           & 542.6 $\pm$ 218.48        \\
 &        & - & Our greedy     & Our greedy & 25.6 $\pm$ 0.15  & 23.77 $\pm$ 0.75 & 7.19 $\pm$ 2.57            & 470.8 $\pm$ 224.58        \\
 &        & - & Our greedy     & ILP        & 25.57 $\pm$ 0.15 & 19.65 $\pm$ 0.47 & 23.14 $\pm$ 1.53           & 466.0 $\pm$ 150.52        \\
 &        & - & ILP            & ILP        & 25.66 $\pm$ 0.09 & 19.66 $\pm$ 0.43 & \textbf{23.38 $\pm$ 1.72}  & 403.2 $\pm$ 74.25         \\
 & TENSAT & 1 & -              & ILP        & 25.6 $\pm$ 0.14  & 21.03 $\pm$ 0.1  & 17.83 $\pm$ 0.29           & \textbf{1.46 $\pm$ 0.04}  \\ \hline
\multirow{6}{*}{NASRNN} &
  \multirow{5}{*}{MCTS} &
  - &
  Default greedy &
  Default greedy &
  1.74 $\pm$ 0.05 &
  1.5 $\pm$ 0.17 &
  13.49 $\pm$ 11.08 &
  127.6 $\pm$ 11.41 \\
 &        & - & Default greedy & ILP        & 1.74 $\pm$ 0.04  & 1.66 $\pm$ 0.04  & 4.89 $\pm$ 1.2             & 122.0 $\pm$ 5.79          \\
 &        & - & Our greedy     & Our greedy & 1.78 $\pm$ 0.04  & 1.57 $\pm$ 0.18  & 11.8 $\pm$ 9.36            & 192.6 $\pm$ 83.35         \\
 &        & - & Our greedy     & ILP        & 1.74 $\pm$ 0.05  & 1.41 $\pm$ 0.23  & \textbf{18.93 $\pm$ 12.03} & 174.0 $\pm$ 64.81         \\
 &        & - & ILP            & ILP        & 1.79 $\pm$ 0.03  & 1.45 $\pm$ 0.26  & 18.74 $\pm$ 13.16          & 1218.4 $\pm$ 1330.71      \\
 & TENSAT & 3 & -              & ILP        & 1.72 $\pm$ 0.03  & 1.55 $\pm$ 0.17  & 10.22 $\pm$ 10.04          & \textbf{12.81 $\pm$ 3.58} \\ \hline
\multirow{6}{*}{ResNet-50} &
  \multirow{5}{*}{MCTS} &
  - &
  Default greedy &
  Default greedy &
  8.67 $\pm$ 0.83 &
  8.67 $\pm$ 0.82 &
  -0.02 $\pm$ 0.12 &
  16.0 $\pm$ 0.71 \\
 &        & - & Default greedy & ILP        & 8.98 $\pm$ 0.85  & 8.95 $\pm$ 0.84  & 0.29 $\pm$ 0.14            & 16.2 $\pm$ 0.84           \\
 &        & - & Our greedy     & Our greedy & 8.73 $\pm$ 0.69  & 8.73 $\pm$ 0.69  & 0.05 $\pm$ 0.1             & 16.0 $\pm$ 1.22           \\
 &        & - & Our greedy     & ILP        & 8.99 $\pm$ 0.8   & 8.92 $\pm$ 0.79  & \textbf{0.78 $\pm$ 0.63}   & 16.8 $\pm$ 0.84           \\
 &        & - & ILP            & ILP        & 8.24 $\pm$ 0.65  & 8.19 $\pm$ 0.67  & 0.66 $\pm$ 0.63            & 16.0 $\pm$ 0.71           \\
 & TENSAT & 4 & -              & ILP        & 8.87 $\pm$ 0.67  & 8.85 $\pm$ 0.65  & 0.31 $\pm$ 0.23            & \textbf{4.72 $\pm$ 0.08}  \\ \hline
\multirow{6}{*}{ResNeXt-50} &
  \multirow{5}{*}{MCTS} &
  - &
  Default greedy &
  Default greedy &
  13.05 $\pm$ 0.29 &
  9.43 $\pm$ 0.18 &
  27.69 $\pm$ 2.67 &
  98.6 $\pm$ 15.21 \\
 &        & - & Default greedy & ILP        & 13.05 $\pm$ 0.3  & 9.4 $\pm$ 0.16   & 27.99 $\pm$ 1.15           & 98.4 $\pm$ 4.39           \\
 &        & - & Our greedy     & Our greedy & 12.85 $\pm$ 0.34 & 9.29 $\pm$ 0.23  & 27.69 $\pm$ 1.06           & 117.6 $\pm$ 12.54         \\
 &        & - & Our greedy     & ILP        & 12.9 $\pm$ 0.47  & 9.22 $\pm$ 0.15  & 28.43 $\pm$ 2.61           & 145.2 $\pm$ 32.39         \\
 &        & - & ILP            & ILP        & 13.04 $\pm$ 0.42 & 9.27 $\pm$ 0.15  & \textbf{28.86 $\pm$ 2.16}  & 296.0 $\pm$ 18.81         \\
 & TENSAT & 4 & -              & ILP        & 13.4 $\pm$ 0.57  & 9.76 $\pm$ 0.2   & 27.11 $\pm$ 3.61           & \textbf{4.82 $\pm$ 0.2}   \\ \hline
\multirow{6}{*}{SqueezeNet} &
  \multirow{5}{*}{MCTS} &
  - &
  Default greedy &
  Default greedy &
  1.72 $\pm$ 0.11 &
  1.22 $\pm$ 0.02 &
  28.89 $\pm$ 4.47 &
  209.2 $\pm$ 129.38 \\
 &        & - & Default greedy & ILP        & 1.65 $\pm$ 0.12  & 1.17 $\pm$ 0.07  & 29.02 $\pm$ 4.13           & 302.6 $\pm$ 73.46         \\
 &        & - & Our greedy     & Our greedy & 1.71 $\pm$ 0.08  & 1.26 $\pm$ 0.06  & 26.53 $\pm$ 3.25           & 233.6 $\pm$ 66.92         \\
 &        & - & Our greedy     & ILP        & 1.68 $\pm$ 0.15  & 1.11 $\pm$ 0.09  & \textbf{33.17 $\pm$ 11.51} & 365.0 $\pm$ 196.87        \\
 &        & - & ILP            & ILP        & 1.79 $\pm$ 0.04  & 1.21 $\pm$ 0.05  & 32.67 $\pm$ 2.23           & 5544.4 $\pm$ 2817.93      \\
 & TENSAT & 3 & -              & ILP        & 1.71 $\pm$ 0.1   & 1.2 $\pm$ 0.01   & 29.74 $\pm$ 4.52           & \textbf{7.47 $\pm$ 2.4}   \\ \hline
\multirow{6}{*}{VGG-19} &
  \multirow{5}{*}{MCTS} &
  - &
  Default greedy &
  Default greedy &
  6.4 $\pm$ 0.45 &
  5.51 $\pm$ 0.11 &
  13.71 $\pm$ 4.11 &
  1.0 $\pm$ 0.0 \\
 &        & - & Default greedy & ILP        & 6.43 $\pm$ 0.33  & 5.47 $\pm$ 0.05  & \textbf{14.78 $\pm$ 4.46}  & 1.0 $\pm$ 0.0             \\
 &        & - & Our greedy     & Our greedy & 6.21 $\pm$ 0.06  & 5.45 $\pm$ 0.0   & 12.3 $\pm$ 0.89            & 1.0 $\pm$ 0.0             \\
 &        & - & Our greedy     & ILP        & 6.28 $\pm$ 0.09  & 5.46 $\pm$ 0.03  & 13.03 $\pm$ 1.45           & 1.0 $\pm$ 0.0             \\
 &        & - & ILP            & ILP        & 6.23 $\pm$ 0.07  & 5.48 $\pm$ 0.05  & 11.96 $\pm$ 1.28           & 2.2 $\pm$ 0.45            \\
 & TENSAT & 1 & -              & ILP        & 6.27 $\pm$ 0.12  & 5.49 $\pm$ 0.07  & 12.48 $\pm$ 0.51           & \textbf{0.28 $\pm$ 0.01}  \\ \hline
\end{tabular}%
}
\end{table}